\documentclass{article} 
\usepackage{iclr2027_conference,times}

\usepackage{amsmath,amsfonts,bm}

\def\Figref#1{Figure~\ref{#1}}

\def\Secref#1{Section~\ref{#1}}

\def\eqref#1{equation~\ref{#1}}
\def\Eqref#1{Equation~\ref{#1}}

\def\1{\bm{1}}

\def\eps{{\epsilon}}

\def\rs{{\textnormal{s}}}
\def\rt{{\textnormal{t}}}

\def\rx{{\textnormal{x}}}

\def\rell{{\textnormal{$\ell$}}}

\def\rvx{{\mathbf{x}}}

\def\vzero{{\bm{0}}}

\def\vmu{{\bm{\mu}}}
\def\vtheta{{\bm{\theta}}}

\def\ve{{\bm{e}}}

\def\vh{{\bm{h}}}

\def\vx{{\bm{x}}}

\def\mE{{\bm{E}}}

\def\mI{{\bm{I}}}

\def\mX{{\bm{X}}}

\DeclareMathAlphabet{\mathsfit}{\encodingdefault}{\sfdefault}{m}{sl}
\SetMathAlphabet{\mathsfit}{bold}{\encodingdefault}{\sfdefault}{bx}{n}

\def\gJ{{\mathcal{J}}}

\def\gL{{\mathcal{L}}}

\def\gN{{\mathcal{N}}}

\def\gP{{\mathcal{P}}}

\def\gS{{\mathcal{S}}}
\def\gT{{\mathcal{T}}}
\def\gU{{\mathcal{U}}}
\def\gV{{\mathcal{V}}}

\def\gX{{\mathcal{X}}}
\def\gY{{\mathcal{Y}}}

\newcommand{\E}{\mathbb{E}}

\newcommand{\R}{\mathbb{R}}

\DeclareMathOperator*{\argmin}{arg\,min}

\usepackage{hyperref}
\usepackage{url}

\usepackage{amsmath, amsthm, amssymb}
\usepackage[most]{tcolorbox}
\usepackage{enumitem}
\usepackage{booktabs}
\usepackage{subcaption}
\usepackage{multicol}
\usepackage{multirow}
\usepackage{algorithm}
\usepackage{algpseudocode}
\algnewcommand{\LineComment}[1]{\State \(\triangleright\) #1}
\usepackage{wrapfig}
\usepackage{float}
\usepackage{longtable,array}
\newcommand{\tran}{^\mathsf{T}} 
\usepackage{diagbox}

\newtcolorbox{mytheorem}[1]{%
    enhanced,
    colback=blue!5!white,       
    colframe=blue!50!black,     
    fonttitle=\bfseries,        
    coltitle=white,             
    title={#1},        
    attach boxed title to top left={yshift=-2mm, xshift=2mm}, 
    boxed title style={colframe=blue!50!black, colback=blue!50!black},
    boxrule=0.5mm,              
    top=4mm,                    
}
\newcommand{\custompar}[1]{\noindent\textbf{#1.\;}}

\def\ours{\textsc{CDMD}} 

\title{CDMD: A Cross-Dataset Mixed-Type Diffusion Model for Tabular Data}

\author{Mohamed Amine Ketata$^1$, Maximilian Schambach$^2$ \& Stephan Günnemann$^1$  \\
$^1$Munich Data Science Institute, Technical University of Munich, Germany \\
$^2$SAP SE, Germany \\
Correspondence to: \texttt{\href{mailto:a.ketata@tum.de}{a.ketata@tum.de}} \\
}

\iclrfinalcopy 
\begin{document}

\maketitle

\begin{abstract}
Generative models for tabular data are typically trained separately for each dataset, limiting knowledge transfer and requiring the storage of many specialized models. In this paper, we introduce \textsc{CDMD}, a tabular diffusion model trained jointly across heterogeneous datasets with different schemas and variable numbers of numerical and categorical features. Unlike existing cross-dataset tabular diffusion models that operate in continuous representation spaces, \textsc{CDMD} defines diffusion directly over the mixed-type feature space and is trained end-to-end. To accommodate heterogeneous categorical domains, we introduce a schema-restricted reverse-process parameterization for masked diffusion models, in which the output space dynamically adapts to each feature’s vocabulary. We then compose numerical and categorical feature-level diffusion processes into a schema-dependent row-level process. A shared schema-aware Transformer denoiser captures dependencies between features and parameterizes the reverse process across varying schemas. On seven real-world datasets, a single jointly trained \textsc{CDMD} achieves the highest average generation quality among strong single-dataset and cross-dataset baselines, while using substantially fewer total parameters than the collection of separately trained models. Furthermore, pre-training on a corpus of 337 datasets improves generation on previously unseen datasets under both limited target data and limited adaptation epochs. These results demonstrate the potential of direct mixed-type diffusion for shared and transferable tabular data generation. Our code is available at \texttt{\href{https://github.com/ketatam/cdmd}{https://github.com/ketatam/cdmd}}.
\end{abstract}

\section{Introduction}
\label{sec:intro}

Synthetic tabular data generation has the potential to support a wide range of applications, from data augmentation and privacy-preserving data sharing to software testing and simulation~\citep{van2023beyond}.
Such applications are particularly relevant in healthcare, finance, and enterprise settings, where real data is often sensitive and publicly available datasets are scarce~\citep{borisov2022deep}.
Recent advances in tabular generative modeling, particularly with diffusion-based models, have improved the statistical fidelity of synthetic data as well as its utility for training downstream predictive models~\citep{shi2025comprehensive,stoian2025survey}.
However, many established tabular generative models are designed for individual datasets, requiring training and storing a separate model for each target dataset. This approach limits opportunities for cross-dataset knowledge transfer and may compromise the quality of generation on small datasets.
\emph{Cross-dataset tabular generative models} offer a promising alternative: a single model trained across multiple datasets can reduce per-dataset training and storage costs while learning reusable patterns that can support generation in limited-data settings and accelerate adaptation to new datasets~\citep{van2024tabular}.
Developing such models, however, is challenging because tabular datasets vary widely in feature semantics, dimensionality, feature types, and distributions.

To address this heterogeneity, existing cross-dataset tabular generative models often represent mixed-type data in continuous spaces and learn shared diffusion models over these representations.
For instance, \Citet{van2024latable} apply Gaussian diffusion to numerical features and to continuous high-dimensional embeddings of categorical features derived from a pre-trained language model, while \citet{lin2025ctsyn} and \citet{han2026breaking} encode rows from different tables into shared latent representations, perform Gaussian diffusion in the latent space, and decode generated representations back to the feature space.
We discuss these approaches in more detail in Appendix~\ref{sec:related_work}, where we also discuss methods that adapt predictive tabular foundation models for synthetic data generation.
Although these methods demonstrate the potential of cross-dataset generative modeling, latent-space approaches introduce additional encoding and decoding stages, which typically require two-stage training and sampling procedures and make generation quality dependent on both latent modeling and reconstruction accuracy~\citep{lin2025ctsyn}.
Meanwhile, \citet{shi2025tabdiff} have demonstrated the effectiveness of mixed-type diffusion for tabular data generation, albeit in the single-dataset setting.
These considerations motivate investigating cross-dataset generative modeling directly in the original mixed-type feature space.

In this work, we introduce a cross-dataset tabular diffusion model that combines continuous and discrete diffusion to model heterogeneous datasets directly in their mixed-type feature spaces.
Our main technical contributions are threefold.
First, we propose a simple reverse-process parametrization for masked diffusion models~\citep{sahoo2024simple} that dynamically adapts to each categorical feature's vocabulary, enabling a shared diffusion formulation across heterogeneous categorical features.
Second, we develop a mixed-type diffusion process that composes feature-level processes according to a dataset's schema, accommodating rows with varying numbers and types of features.
Third, based on the resulting diffusion objective, we design a Transformer-based denoising network that maps noisy variable-sized mixed-type rows to corresponding predictions that parametrize the reverse denoising process.
We call our model \ours{} (\underline{C}ross-\underline{D}ataset \underline{M}ixed-type \underline{D}iffusion model).

We evaluate \ours{} in two main settings: in-domain generation and adaptation to unseen datasets.
For in-domain generation, we train \ours{} jointly on seven widely used real-world datasets spanning diverse domains.
A single \ours{} model achieves higher average synthetic data quality than state-of-the-art single-dataset baselines, while using a fraction of their combined parameter count.
For adaptation to unseen datasets, we pre-train \ours{} on a corpus of 337 datasets, and evaluate adaptation with either a limited number of training samples or training epochs. Under both adaptation constraints, \ours{} achieves higher generation quality than the baselines.


\section{Problem formulation}
\label{sec:problem_formulation}

\custompar{Notation}
We denote a corpus of $L$ tabular datasets as $\{ \gT^{(\ell)} \}_{\ell=1}^L$, where each dataset $\gT^{(\ell)} = \left(\gS^{(\ell)}, \mX^{(\ell)}\right)$ consists of a table schema $\gS^{(\ell)}$ and a set of rows $\mX^{(\ell)}=\{ \vx_i^{(\ell)} \}_{i=1}^{N^{(\ell)}}$.
Following \Citet{van2024latable}, we define the schema as $\gS^{(\ell)} = \left(C^{(\ell)}, \{ s_j^{(\ell)} \}_{j=1}^{d^{(\ell)}}\right)$, where $C^{(\ell)}$ is the dataset's name or description, $d^{(\ell)}$ is the number of columns (features), and $s_j^{(\ell)}$ is a column schema defining its name, type (numerical or categorical) and domain as follows:
\begin{equation*}
    s_j^{(\ell)} = \big( \underbrace{c_j^{(\ell)}}_{\text{name}}, \underbrace{\tau_j^{(\ell)}}_{\text{type}}, \underbrace{\gX_j^{(\ell)}}_{\text{domain}}\big),
    \text{ where }
    \tau_j^{(\ell)} \in \{ \mathrm{num}, \mathrm{cat} \}
    \text{ and }
    \gX_j^{(\ell)} =
    \begin{cases}
        \R & \text{if } \tau_j^{(\ell)}=\mathrm{num},\\
        \gV_j^{(\ell)} & \text{if } \tau_j^{(\ell)}=\mathrm{cat},
    \end{cases}
\end{equation*}
with $\gV_j^{(\ell)}=\big\{ v_{j,k}^{(\ell)} \big\}_{k=1}^{K_j^{(\ell)}}$ denoting the vocabulary of categorical column $j$ with $K_j^{(\ell)}$ distinct possible values.
Following common practice in tabular data generation~\citep{kotelnikov2023tabddpm}, we assume that each numerical feature is normalized using a transformation fitted only to the corresponding column of the training data.
We use $\gJ_\mathrm{num}^{(\ell)}$ and $\gJ_\mathrm{cat}^{(\ell)}$ to refer to the index sets of numerical and categorical columns, respectively, and $\vx^\mathrm{num}$ and $\vx^\mathrm{cat}$ to refer to the numerical and categorical parts, respectively, of a row $\vx$. 
We additionally define the row space induced by schema $\gS^{(\ell)}$ as $\gX_{\gS^{(\ell)}}=\prod_{j=1}^{d^{(\ell)}}\gX_j^{(\ell)}$.
Throughout the paper, we will drop any index that is irrelevant or clear from context.

\custompar{Tabular generative modeling}
In the context of generative modeling for tabular data, we assume the rows of a given dataset $\gT^{(\ell)}$ to be independent and identically distributed (iid) samples from an unknown data distribution $p_{\mathrm{data}}^{(\ell)}$, i.e., $\{ \vx_i^{(\ell)} \}_{i=1}^{N^{(\ell)}} \overset{\mathrm{iid}}{\sim} p_{\mathrm{data}}^{(\ell)}$, and we aim to approximate it with a parametric distribution $p_\vtheta \approx p_{\mathrm{data}}^{(\ell)}$ that enables sampling synthetic rows.
Given $L$ datasets, \textit{single-dataset generative models} learn a separate parametric distribution for each dataset, $p_{\vtheta^{(\ell)}} \approx p_{\mathrm{data}}^{(\ell)}$ \citep{xu2019modeling, kotelnikov2023tabddpm,shi2025tabdiff}. Consequently, they often hard-code the schema information in the model architecture and parameters.
For instance, they learn column-specific embedding and output layers, and the network width is controlled by the number of columns. This rigidity requires training a new model, or parts of it, from scratch on each dataset.

In contrast, \textit{cross-dataset generative models} learn a single parametric distribution shared between datasets. One approach is by conditioning the distribution of a tabular dataset on its schema, $p_\vtheta\left(\cdot \mid \gS^{(\ell)}\right) \approx p_{\mathrm{data}}^{(\ell)}$.
Formally, $p_\vtheta$ defines a family of probability distributions, mapping a schema $\gS$ to a probability distribution over its row space $\gX_\gS$.
Another approach to cross-dataset modeling is to provide sample rows from the target dataset as context to the model, as done by~\citet{han2026breaking}. A key difference between these two approaches is that in-context generators require target rows during sampling, whereas schema-conditional models can sample from a target dataset given only its schema. We focus on schema-conditional approaches in this work.
To the best of our knowledge, existing cross-dataset tabular diffusion models (both in-context and schema-conditional) define their learned distributions over a continuous space shared between features and datasets~\citep{van2024latable,lin2025ctsyn,han2026breaking}.
In this work, we propose a cross-dataset schema-conditional \emph{mixed-type} diffusion model that leverages both continuous and discrete diffusion.

\section{Cross-Dataset Mixed-type Diffusion model (CDMD)}
\label{sec:method}

\custompar{Overview}
To address the heterogeneity across datasets without relying on a shared continuous representation, we propose a fine-to-coarse approach that dynamically combines feature-level diffusion processes into a row-level process based on the target schema. Our approach can be summarized as follows:
\textbf{(1)} For each data type, numerical or categorical, we define an atomic one-dimensional diffusion process that can model an arbitrary tabular feature of that type (\Secref{sec:feature_level_diffusion}). We devote special care to addressing the heterogeneity across discrete vocabularies of categorical features.
\textbf{(2)} Given a table schema $\gS$, we compose these one-dimensional diffusion processes to \emph{dynamically} define a $d$-dimensional diffusion process to model tabular rows (\Secref{sec:row_level_diffusion}).
\textbf{(3)} We parametrize the row-level reverse diffusion process using a schema-conditional, transformer-based denoising architecture (\Secref{sec:model_architecture},~\Figref{fig:overview}).


\subsection{Feature-level diffusion processes}
\label{sec:feature_level_diffusion}

The first step towards our goal is to develop feature-level diffusion processes that can model any numerical or categorical feature and allow us to train a shared schema-conditional neural network on heterogeneous columns.
In this section, $\rx$ denotes a one-dimensional random variable representing one tabular feature with column schema $s$.
The main idea behind diffusion models is to define a noising process that progressively corrupts the observed data and to learn the reverse denoising process, thereby enabling the sampling of new data points from pure noise \citep{sohl2015deep}.
Let $t\in [0,1]$ be the diffusion time and $\rx_t\sim p_t$ the distribution of $\rx$ at time $t$.
Specifically, $\rx_0 \sim p_0(\cdot\mid s)$ is distributed according to the data distribution, and $\rx_1 \sim p_1$ is sampled from the prior distribution.
We use $q$ to refer to the diffusion kernel controlling the state transitions between two arbitrary time steps $t,t'\in[0,1]$ such that $p_t(x_t)=\E_{\rx_{t'}\sim p_{t'}}[q(x_t\mid x_{t'})]$.

\custompar{Gaussian diffusion for normalized numerical features}
Numerical features share the same domain, $\R$; therefore, we can model their distributions with a shared noising process and a schema-conditional reverse process.
Following prior work on tabular data generation~\citep{zhang2024mixed,shi2025tabdiff}, we adopt the Variance Exploding formulation of score-based generative models~\citep{song2020score} whose noising process corresponds to the Gaussian diffusion kernel
\begin{equation}
\label{eq:gaussian_forward_diffusion}
    q_\mathrm{num}\left(x_t \mid x_0\right) = \gN\left(x_t;x_0,\sigma_t^2\right),
\end{equation}
where $\sigma_t\colon[0,1]\to\R_+$ is the Gaussian diffusion schedule.
To model the reverse process, we parametrize the score function of $p_t$ as $\frac{\partial}{\partial x_t} \log p_t(x_t) \approx\frac{\mu_\vtheta(x_t,t,s)-x_t}{\sigma_t^2}$~\citep{karras2022elucidating}, where $\mu_\vtheta(\cdot,\cdot,s)\colon\R\times[0,1]\to\R$ is a schema-conditional denoiser network that predicts clean data from noisy data. To train $\mu_\vtheta$ on multiple features, we minimize the expected $L_2$ denoising loss,
\begin{equation}
\label{eq:numerical_loss}
    \gL_\mathrm{num}(\vtheta) = \E_{\rs\sim p_s} \E_{\rx_0\sim p_0(\cdot\mid s)}\E_{\rt\sim \gU([0,1])}\E_{\rx_t\sim q_\mathrm{num}(\cdot\mid x_0)} \left[\lambda_t\left(\mu_\vtheta(x_t,t,s)-x_0\right)^2\right],
\end{equation}
where $p_s$ is a distribution over the features available for training and $\lambda_t$ is a time-dependent weighting function \citep{karras2022elucidating}. Once trained, we can sample from the learned distribution corresponding to any target schema $s$ by numerically solving the (schema-conditioned) flow ordinary differential equation (ODE) \citep{song2020score}, with $\sigma_t'$ denoting the time derivative of $\sigma_t$,
\begin{equation}
\label{eq:reverse_ode}
    dx=- \sigma_t' / \sigma_t \left( \mu_\vtheta(x,t,s) - x \right)  dt.
\end{equation} 

\custompar{Generalized masked diffusion for categorical features}
In contrast to numerical features, which share the same domain, $\R$, and to discrete data from other domains, such as language modeling, where tokens share a common vocabulary, tabular categorical features usually have distinct (possibly overlapping) domains.
To address this heterogeneity, we develop a generalized discrete diffusion process defined over the joint space of all vocabularies of interest, which can be efficiently applied to an arbitrary number of categorical features thanks to a novel reverse-process parametrization.

In this work, we adapt masked diffusion models \citep{austin2021structured,sahoo2024simple} to our setting for their simplicity and strong empirical performance in tabular data generation~\citep{shi2025tabdiff}.
Formally, our generalized process is defined over $\gU$, the universe of all categories from our datasets of interest, augmented with a special \textsc{[MASK]} category denoted by $m$, and we demonstrate how it can be applied to efficiently model an arbitrary categorical random variable $\rx$ with schema $s$ and domain $\gV \subset \gU$.
In masked diffusion models, the noising process is an absorbing process that sets the clean value $x_0$ to the masked state $m$ with probability $1-\alpha_t$ by time $t$, where $\alpha_t\colon [0,1]\to[0,1]$ is the masked diffusion schedule. The masked diffusion kernel can be written as \citep{sahoo2024simple}
\begin{equation}
\label{eq:masked_forward_diffusion}
    q_\mathrm{cat}(x_t\mid x_0) = \alpha_t \1_{x_t=x_0} + (1-\alpha_t) \1_{x_t=m},
\end{equation}
where $\1_\mathrm{condition}$ is the indicator function, evaluating to $1$ if $\mathrm{condition}$ is true, and $0$ otherwise.
This formulation implies that for all $t\in[0,1]$, $x_t$ can either be $m$ or $x_0$.
The true posterior distribution from $t$ to $t'<t$, conditioned on $x_0$, can be written according to these two cases \citep{sahoo2024simple}:
\begin{equation}
\label{eq:true_masked_reverse_process}
    q_\mathrm{cat}(x_{t'}\mid x_t,x_0) =
    \begin{cases}
        \1_{x_{t'}=x_t}, & \text{ if } x_t\neq m, \\
        \frac{\alpha_{t'}-\alpha_t}{1-\alpha_t}\1_{x_{t'}=x_0} + \frac{1-\alpha_{t'}}{1-\alpha_t}\1_{x_{t'}=m}, & \text{ if } x_t=m.
    \end{cases}
\end{equation}
Intuitively, in the reverse process, when $x_t$ is masked, the clean value is recovered with probability $\frac{\alpha_{t'}-\alpha_t}{1-\alpha_t}$ before time $t'$, and once unmasked, the value remains fixed until the end ($t=0$).
To learn the reverse unmasking process, \citet{sahoo2024simple} propose a reverse process parametrization that matches the form of~\Eqref{eq:true_masked_reverse_process} but replaces (the unknown) $x_0$ by a network prediction. Importantly, to make this process scalable to a large number of features, we additionally leverage the fact that $x_0\in\gV$ to constrain the network's predictions to the valid categories. 
Formally, with $\Delta^{K-1}=\{ \bm{\pi} \in [0,1]^{K} \mid \sum_{i=1}^{K} \pi_i = 1 \}$ denoting the ($K-1$)-simplex, our network $\vmu_\vtheta(\cdot,\cdot,s)\colon\gU\times[0,1]\to\Delta^{|\gV|-1}$ is a schema-conditional classifier over the categories in $\gV$.
Using the notation $\bm{\pi}[v]$ to denote the probability assigned by $\bm{\pi} \in \Delta^{|\gV|-1}$ to category $v\in\gV$, our proposed reverse-process parametrization is given by
\begin{equation}
\label{eq:parametrized_masked_reverse_process}
    p_\vtheta(x_{t'}\mid x_t,s) = 
    \begin{cases}
        \1_{x_{t'}=x_t}, & \text{ if } x_t\neq m, \\
        \frac{\alpha_{t'}-\alpha_t}{1-\alpha_t} \vmu_\vtheta(x_t,t,s)[x_{t'}]\1_{x_{t'}\in\gV} + \frac{1-\alpha_{t'}}{1-\alpha_t}\1_{x_{t'}=m}, & \text{ if } x_t=m.
    \end{cases}
\end{equation}
Our parametrization can be viewed as a generalization of the SUBS parametrization of \citet{sahoo2024simple}, which can be recovered by setting $\gV=\gU\setminus \{m\}$, corresponding to the case where all categorical features share the same domain. Crucially, to evaluate~\Eqref{eq:parametrized_masked_reverse_process}, our model needs only to predict probabilities for categories in $\gV$ whose size is independent of $\gU$ and typically $|\gV| \ll |\gU|$. This is exactly what makes our process scalable to an arbitrary number of features. Moreover, because our parametrization is independent of the exact definition of $\gU$, it natively supports settings in which $\gU$ is dynamically updated, such as adapting the model to new categorical distributions.
Since we use the same noising process as \citet{sahoo2024simple} and our reverse-process parametrization retains the same analytical form as theirs, we can directly use their simplified likelihood bound, which is a weighted cross-entropy loss, to train our model across heterogeneous categorical features,
\begin{equation}
\label{eq:masked_loss}
    \gL_\mathrm{cat}(\vtheta) = \E_{\rs\sim p_s} \E_{\rx_0\sim p_0(\cdot\mid s)} \E_{\rt\sim \gU([0,1])}\E_{\rx_t\sim q_\mathrm{cat}(\cdot\mid x_0)} \left[ \1_{x_t=m} \frac{\alpha_t'}{1-\alpha_t}\log \left( \vmu_\vtheta\left(x_t,t,s\right)\left[x_0\right] \right) \right].
\end{equation}

\subsection{Row-level diffusion process}
\label{sec:row_level_diffusion}

Having defined one-dimensional diffusion processes for numerical and categorical features separately, we can now use them as building blocks to define a row-level mixed-type diffusion process dynamically for any table schema.
Here, $\rvx$ denotes a $d$-dimensional mixed-type random variable representing the row distribution of a table with schema $\gS$.
We apply the row-level noising process by applying the feature-level processes independently to each feature, analogously to how pixels are noised independently in image diffusion models~\citep{song2020score} and how tokens are masked independently in diffusion language models~\citep{sahoo2024simple}.
Since the choice of the feature-level process depends on the feature type, the row-level noising process is conditioned on the schema:
\begin{equation}
    q\left(\vx_t\mid\vx_0,\gS\right) = \prod\nolimits_{j\in\gJ_\mathrm{num}} q_\mathrm{num}\left(x_{t,j}\mid x_{0,j}\right) \prod\nolimits_{j\in\gJ_\mathrm{cat}}q_\mathrm{cat}\left(x_{t,j}\mid x_{0,j}\right),
\end{equation}
where $q_\mathrm{num}$ and $q_\mathrm{cat}$ are the feature-level noising processes defined in Equations \ref{eq:gaussian_forward_diffusion} and \ref{eq:masked_forward_diffusion}, respectively.
To model the row-level reverse process, the network needs to predict the clean values for the numerical features, and the unmasked categories for the (masked) categorical features.
Crucially, by processing the full row jointly, the model can capture row-level distributions beyond column marginals.
Formally, our final network $\vmu_\vtheta(\cdot,\cdot,\gS)\colon \left(\prod_{j=1}^d \gX_j\right) \times [0,1] \to \prod_{j=1}^d\gY_j$, with
\begin{equation}
    \gX_j = 
    \begin{cases} 
    \R, &\text{if } j\in\gJ_\mathrm{num}, \\
    \gU, &\text{if } j\in\gJ_\mathrm{cat},
    \end{cases}
    \qquad
    \text{and} 
    \qquad
    \gY_j = 
    \begin{cases} 
    \R, & \text{if } j\in\gJ_\mathrm{num}, \\
    \Delta^{|\gV_j|-1}, & \text{if } j\in\gJ_\mathrm{cat},
    \end{cases}
\end{equation}
is a schema-conditional row-level denoiser with variable-sized input and output.
We can now finally train this network end-to-end on a collection of $L$ tabular datasets via the combined mixed-type loss,
\begin{multline}
\label{eq:total_loss}
    \gL(\vtheta) = \E_{\rell\sim p(\{1,2,\dots,L \})} \E_{\rvx_0\sim p_\mathrm{data}^{(\ell)}}\E_{\rt\sim \gU([0,1])}\E_{\rvx_t \sim q(\cdot\mid \vx_0)} \bigg[\frac{\lambda_t}{|\gJ_\mathrm{num}|} \left\|\vmu_{\vtheta}^\mathrm{num}\left(\vx_t,t,\gS\right) - \vx_0^\mathrm{num} \right\|_2^2 \\
     \qquad + \frac{\alpha_t'}{|\gJ_\mathrm{cat}|(1-\alpha_t)} \sum_{j\in\gJ_\mathrm{cat},x_{t,j}=m} \log \left( \vmu_{\vtheta,j}\left(\vx_t,t,\gS\right)[x_{0,j}] \right)   \bigg].
\end{multline}

\custompar{Sampling}
Once trained, we can sample synthetic rows from our model for a target dataset given its schema.
First, we initialize numerical features with Gaussian noise at the terminal noise level and categorical features with the mask state. 
We then iteratively reverse the diffusion process along a discretized time grid. At each step, the trained denoiser predicts clean numerical values and categorical probabilities from the current noisy row, conditioned on the schema and time. 
These predictions determine an Euler step for the numerical probability-flow ODE (\Eqref{eq:reverse_ode}) and the categorical reverse transitions (\Eqref{eq:parametrized_masked_reverse_process}) from which we sample. 
We adopt the mixed-type stochastic sampler of \textsc{TabDiff}~\citep{shi2025tabdiff}, based on the \textsc{EDM} sampler~\citep{karras2022elucidating}, which injects additional Gaussian noise and probabilistically remasks categorical features before each denoising update. Finally, we invert the numerical preprocessing to recover the original feature scales.

We provide additional details on the diffusion model in Appendix~\ref{app:diffusion_details} and pseudocode for the training and sampling procedure in Algorithms~\ref{alg:training} and~\ref{alg:sampling}, respectively.


\subsection{Schema-conditional denoiser network architecture}
\label{sec:model_architecture}

So far, we have developed a framework for training a single diffusion model on heterogeneous tabular datasets by parametrizing the reverse process via a schema-conditional denoiser network.
From the diffusion process perspective, the denoiser, $\vmu_\vtheta$, needs to have the following key properties:
(\textit{i}) It takes as input a \textit{variable-sized} \textit{mixed-type} (noisy) row and a corresponding schema.
(\textit{ii}) By leveraging schema information, it predicts a scalar value for every numerical feature and a categorical distribution over the valid categories for every categorical feature.
(\textit{iii}) The output is equivariant to the joint permutation of the input row and schema (see Appendix~\ref{app:desiderata} for a detailed discussion on permutation equivariance).
Guided by these properties, we propose a concrete architecture for $\vmu_\vtheta$ consisting of a tokenizer, a transformer encoder, and a detokenizer, visualized in \Figref{fig:overview}.
The tokenizer represents each noisy feature as a token, the encoder models cross-feature dependencies, and the detokenizer produces numerical value estimates and categorical probability vectors. All parameters are shared across datasets, and type-specific modules are shared between all columns of that type.
Let $D$ denote the token dimension.

\custompar{Schema Embeddings}
Most schema information is typically available in textual form. To obtain a numerical representation of this information, we use a frozen pre-trained text embedding model $f_\mathrm{text}$ that maps strings to $d_\mathrm{emb}$-dimensional vectors. We follow \citet{lin2025ctsyn} and use \textsc{GTE}~\citep{li2023towards} as the text embedding model.
We apply $f_\mathrm{text}$ to the table's name, column names, and categories, followed by learned linear projection layers to obtain $D$-dimensional schema embeddings corresponding to each textual schema component
\begin{equation*}
    \ve_\mathrm{name}, \qquad
    \left\{\ve_{\mathrm{col},j} \right\}_{j=1}^d, \qquad
    \left\{\mE_{\mathrm{cat},j}=[\ve_{\mathrm{cat},j,1},\dots,\ve_{\mathrm{cat},j,K_j}]\tran \right\}_{j=1}^d.
\end{equation*}


\custompar{Tokenizer}
For each noisy feature $x_{t,j}$, we compute a value embedding $\ve_{\mathrm{val},j} \in\R^D$, depending on its type and whether it is masked or not
\begin{equation}
    \ve_{\mathrm{val},j} =
    \begin{cases}
        \mathrm{MLP}_\mathrm{num,in}(x_{t,j}), & \text{if } j\in\gJ_\mathrm{num}, \\
        \ve_{\mathrm{cat},j,k}, & \text{if } j\in\gJ_\mathrm{cat} \text{ and } x_{t,j} = v_{j,k}, \\
        \mathrm{Attention}\left(\ve_m\tran, \mE_{\mathrm{cat},j},\mE_{\mathrm{cat},j}\right)\tran, & \text{if } j\in\gJ_\mathrm{cat} \text{ and } x_{t,j} = m.
    \end{cases}
\end{equation}
To obtain the value embeddings, numerical features are processed by an MLP, while unmasked categorical features use their category embeddings. For a masked categorical feature, a learned mask embedding $\ve_m\in\R^D$ attends to the category embeddings in $\mE_{\mathrm{cat},j}$, which serve as both keys and values. While $\ve_m$ is shared across categorical columns and datasets, the attention mechanism's output depends on the current column's vocabulary.
We then add type, column, table, and time embeddings to the value embedding to obtain the feature token, $\vh_j$:
\begin{equation}
    \vh_j = \ve_{\mathrm{val},j} + \ve_{\tau_j} + \ve_{\mathrm{col},j} + \ve_\mathrm{name} + \ve_\mathrm{time},
\end{equation}
where $\ve_{\tau_j} \in \{\ve_\mathrm{num},\ve_\mathrm{cat}\}$ are learnable type embeddings and $\ve_\mathrm{time}$ is a sinusoidal embedding of the time step $t$ followed by an MLP \citep{kotelnikov2023tabddpm}.

\custompar{Backbone}
A transformer encoder \citep{vaswani2017attention} processes the feature tokens jointly to capture cross-column relationships
\begin{equation}
    \left[\hat\vh_1,\dots,\hat\vh_d\right]\tran = \mathrm{TransformerEncoder}\left(\left[\vh_1,\dots,\vh_d\right]\tran\right) \in\R^{d\times D} .
\end{equation}
The encoder does \textit{not} use positional encodings. Together with the feature-wise tokenizer and detokenizer, this makes the model equivariant to jointly permuting columns and their corresponding schema entries. Column identities are preserved through the schema embeddings.

\custompar{Detokenizer}
The detokenizer maps the processed token for column $j$, $\hat\vh_j$, to the final output. For numerical columns, a shared MLP predicts the clean feature value
\begin{equation}
    \vmu_{\vtheta,j}(\vx_t,t,\gS) = \mathrm{MLP}_\mathrm{num,out}(\hat\vh_j) \in \R, \quad j\in\gJ_\mathrm{num}.
\end{equation}
For categorical columns, a shared single-layer transformer decoder takes the $K_j$ category embeddings as target tokens and the contextualized feature as a single memory token
\begin{equation}
    \left[\hat\ve_{\mathrm{cat},j,1},\dots,\hat\ve_{\mathrm{cat},j,K_j}\right]\tran = \mathrm{TransformerDecoder}\left(\mathrm{target}=\mE_{\mathrm{cat},j}, \mathrm{memory}=\hat\vh_j\tran\right) \in\R^{K_j\times D}
\end{equation}
Then, a shared scalar projection produces category logits, which are fed into a softmax function,
\begin{equation}
    \left\{ z_{j,k} = \mathrm{Lin}_\mathrm{out}(\hat\ve_{\mathrm{cat},j,k})\right\}_{k=1}^{K_j}, \qquad \vmu_{\vtheta,j}(\vx_t,t,\gS) = \mathrm{Softmax}\left(\left\{ z_{j,k} \right\}_{k=1}^{K_j}\right), \quad j\in\gJ_\mathrm{cat}.
\end{equation}


\section{Experiments}
\label{sec:experiments}

\begin{table}[t]
\caption{In-domain cross-dataset generation results (\Secref{sec:exp:cross_dataset_generation}). Entries report mean $\scriptstyle\pm$ standard deviation over three runs. Average is the mean over datasets. \textsc{Train Set} is the real-data reference. Bold and underlined entries mark the best and second-best scores in each column, excluding \textsc{Train Set}. \textsc{GReaT} cannot be applied to News because of its maximum length limit.}
\label{tab:cross-table-results}
\centering
\small
\resizebox{\linewidth}{!}{%
\begin{tabular}{@{}c@{\hspace{0.4em}}l|c|c|c|c|c|c|c|c}
\toprule
\multicolumn{10}{c}{Generation Quality: $Q=(F+U)/2\;$ ($\uparrow$)} \\
\midrule
 & {\renewcommand{\arraystretch}{0.6}\diagbox[font={\fontsize{5.5}{6.6}\selectfont},width=8.5em,height=1.1em]{Method}{Dataset}} & Adult & Default & Magic & Shoppers & Diabetes & Beijing & News & Average \\
\midrule
 & \textsc{Train Set} & \(89.64\,{\scriptstyle \pm 0.00}\) & \(82.52\,{\scriptstyle \pm 0.00}\) & \(89.69\,{\scriptstyle \pm 0.00}\) & \(89.57\,{\scriptstyle \pm 0.00}\) & \(77.40\,{\scriptstyle \pm 0.00}\) & \(88.15\,{\scriptstyle \pm 0.00}\) & \(87.47\,{\scriptstyle \pm 0.00}\) & \(86.35\) \\
\midrule[0.2pt]
\multirow{7}{*}[0pt]{\rotatebox[origin=c]{90}{\fontsize{5.5}{6.6}\selectfont Single-dataset}} & \textsc{SMOTE} & \(86.17\,{\scriptstyle \pm 0.03}\) & \(79.33\,{\scriptstyle \pm 0.05}\) & \(88.40\,{\scriptstyle \pm 0.12}\) & \(86.73\,{\scriptstyle \pm 0.06}\) & \(72.41\,{\scriptstyle \pm 0.05}\) & \(\mathbf{86.72}\,{\scriptstyle \pm 0.03}\) & \(\underline{85.66}\,{\scriptstyle \pm 0.38}\) & \(83.63\) \\
 & \textsc{CTGAN} & \(78.06\,{\scriptstyle \pm 1.43}\) & \(68.44\,{\scriptstyle \pm 0.41}\) & \(74.56\,{\scriptstyle \pm 1.35}\) & \(74.69\,{\scriptstyle \pm 2.00}\) & \(59.96\,{\scriptstyle \pm 2.40}\) & \(81.28\,{\scriptstyle \pm 0.51}\) & \(77.64\,{\scriptstyle \pm 0.38}\) & \(73.52\) \\
 & \textsc{TVAE} & \(82.47\,{\scriptstyle \pm 0.84}\) & \(74.15\,{\scriptstyle \pm 0.56}\) & \(80.26\,{\scriptstyle \pm 1.37}\) & \(72.92\,{\scriptstyle \pm 0.50}\) & \(60.53\,{\scriptstyle \pm 1.27}\) & \(81.61\,{\scriptstyle \pm 0.96}\) & \(80.45\,{\scriptstyle \pm 0.27}\) & \(76.06\) \\
 & \textsc{GReaT} & \(82.64\,{\scriptstyle \pm 0.21}\) & \(76.54\,{\scriptstyle \pm 0.36}\) & \(82.04\,{\scriptstyle \pm 0.13}\) & \(83.46\,{\scriptstyle \pm 0.18}\) & \(70.32\,{\scriptstyle \pm 0.14}\) & \(84.14\,{\scriptstyle \pm 0.09}\) & -- & \(79.86\) \\
 & \textsc{TabDDPM} & \(87.76\,{\scriptstyle \pm 0.04}\) & \(79.85\,{\scriptstyle \pm 0.38}\) & \(88.33\,{\scriptstyle \pm 0.10}\) & \(85.40\,{\scriptstyle \pm 0.12}\) & \(41.48\,{\scriptstyle \pm 0.18}\) & \(86.00\,{\scriptstyle \pm 0.31}\) & \(38.59\,{\scriptstyle \pm 5.89}\) & \(72.49\) \\
 & \textsc{TabSyn} & \(\underline{88.43}\,{\scriptstyle \pm 0.07}\) & \(81.36\,{\scriptstyle \pm 0.12}\) & \(\underline{88.85}\,{\scriptstyle \pm 0.00}\) & \(87.84\,{\scriptstyle \pm 0.30}\) & \(72.33\,{\scriptstyle \pm 0.08}\) & \(\underline{86.18}\,{\scriptstyle \pm 0.11}\) & \(84.87\,{\scriptstyle \pm 0.16}\) & \(\underline{84.27}\) \\
 & \textsc{TabDiff} & \(88.28\,{\scriptstyle \pm 0.10}\) & \(\underline{81.39}\,{\scriptstyle \pm 0.23}\) & \(88.67\,{\scriptstyle \pm 0.06}\) & \(\underline{87.93}\,{\scriptstyle \pm 0.25}\) & \(\underline{75.38}\,{\scriptstyle \pm 0.17}\) & \(86.15\,{\scriptstyle \pm 0.30}\) & \(81.75\,{\scriptstyle \pm 0.11}\) & \(84.22\) \\
\midrule[0.2pt]
\multirow{3}{*}[0pt]{\rotatebox[origin=c]{90}{\fontsize{5.5}{6.6}\selectfont Cross-dataset}} & \textsc{TabEBM} & \(62.40\,{\scriptstyle \pm 0.60}\) & \(57.41\,{\scriptstyle \pm 0.96}\) & \(84.25\,{\scriptstyle \pm 0.17}\) & \(70.00\,{\scriptstyle \pm 0.56}\) & \(43.46\,{\scriptstyle \pm 0.37}\) & \(77.31\,{\scriptstyle \pm 0.11}\) & \(67.66\,{\scriptstyle \pm 5.85}\) & \(66.07\) \\
 & \textsc{CTSyn} & \(67.66\,{\scriptstyle \pm 0.62}\) & \(66.80\,{\scriptstyle \pm 0.79}\) & \(72.46\,{\scriptstyle \pm 0.35}\) & \(63.67\,{\scriptstyle \pm 1.71}\) & \(48.04\,{\scriptstyle \pm 0.04}\) & \(76.07\,{\scriptstyle \pm 0.05}\) & \(79.66\,{\scriptstyle \pm 0.09}\) & \(67.76\) \\
 & \textsc{CDMD (Ours)} & \(\mathbf{88.59}\,{\scriptstyle \pm 0.03}\) & \(\mathbf{81.68}\,{\scriptstyle \pm 0.07}\) & \(\mathbf{88.92}\,{\scriptstyle \pm 0.08}\) & \(\mathbf{88.48}\,{\scriptstyle \pm 0.05}\) & \(\mathbf{76.09}\,{\scriptstyle \pm 0.05}\) & \(85.90\,{\scriptstyle \pm 0.27}\) & \(\mathbf{86.00}\,{\scriptstyle \pm 0.03}\) & \(\mathbf{85.09}\) \\
\midrule[\heavyrulewidth]
\multicolumn{10}{c}{Overall Evaluation Score: $S=(F+U+P)/3\;$ ($\uparrow$)} \\
\midrule
 & {\renewcommand{\arraystretch}{0.6}\diagbox[font={\fontsize{5.5}{6.6}\selectfont},width=8.5em,height=1.1em]{Method}{Dataset}} & Adult & Default & Magic & Shoppers & Diabetes & Beijing & News & Average \\
\midrule
 & \textsc{Train Set} & \(59.79\,{\scriptstyle \pm 0.00}\) & \(55.05\,{\scriptstyle \pm 0.00}\) & \(59.99\,{\scriptstyle \pm 0.00}\) & \(60.09\,{\scriptstyle \pm 0.00}\) & \(51.60\,{\scriptstyle \pm 0.00}\) & \(58.77\,{\scriptstyle \pm 0.00}\) & \(58.31\,{\scriptstyle \pm 0.00}\) & \(57.66\) \\
\midrule[0.2pt]
\multirow{7}{*}[0pt]{\rotatebox[origin=c]{90}{\fontsize{5.5}{6.6}\selectfont Single-dataset}} & \textsc{SMOTE} & \(65.30\,{\scriptstyle \pm 0.04}\) & \(60.79\,{\scriptstyle \pm 0.03}\) & \(64.60\,{\scriptstyle \pm 0.10}\) & \(63.46\,{\scriptstyle \pm 0.18}\) & \(52.47\,{\scriptstyle \pm 0.01}\) & \(66.62\,{\scriptstyle \pm 0.04}\) & \(61.89\,{\scriptstyle \pm 0.22}\) & \(62.16\) \\
 & \textsc{CTGAN} & \(78.34\,{\scriptstyle \pm 0.87}\) & \(72.21\,{\scriptstyle \pm 0.35}\) & \(78.93\,{\scriptstyle \pm 0.93}\) & \(76.57\,{\scriptstyle \pm 1.38}\) & \(68.88\,{\scriptstyle \pm 1.27}\) & \(82.85\,{\scriptstyle \pm 0.29}\) & \(80.26\,{\scriptstyle \pm 0.33}\) & \(76.86\) \\
 & \textsc{TVAE} & \(80.80\,{\scriptstyle \pm 0.40}\) & \(76.69\,{\scriptstyle \pm 0.05}\) & \(80.64\,{\scriptstyle \pm 0.53}\) & \(77.63\,{\scriptstyle \pm 0.39}\) & \(67.80\,{\scriptstyle \pm 0.69}\) & \(82.60\,{\scriptstyle \pm 0.41}\) & \(80.73\,{\scriptstyle \pm 0.12}\) & \(78.13\) \\
 & \textsc{GReaT} & \(79.49\,{\scriptstyle \pm 0.19}\) & \(76.07\,{\scriptstyle \pm 0.39}\) & \(80.31\,{\scriptstyle \pm 0.14}\) & \(80.98\,{\scriptstyle \pm 0.28}\) & \(71.62\,{\scriptstyle \pm 0.07}\) & \(80.51\,{\scriptstyle \pm 0.05}\) & -- & \(78.16\) \\
 & \textsc{TabDDPM} & \(83.34\,{\scriptstyle \pm 0.11}\) & \(78.13\,{\scriptstyle \pm 0.26}\) & \(83.18\,{\scriptstyle \pm 0.18}\) & \(78.64\,{\scriptstyle \pm 0.18}\) & \(60.53\,{\scriptstyle \pm 0.08}\) & \(\mathbf{84.43}\,{\scriptstyle \pm 0.25}\) & \(59.06\,{\scriptstyle \pm 3.93}\) & \(75.33\) \\
 & \textsc{TabSyn} & \(\underline{83.61}\,{\scriptstyle \pm 0.07}\) & \(\underline{79.00}\,{\scriptstyle \pm 0.02}\) & \(\underline{83.99}\,{\scriptstyle \pm 0.10}\) & \(\underline{83.30}\,{\scriptstyle \pm 0.41}\) & \(\underline{74.38}\,{\scriptstyle \pm 0.09}\) & \(\underline{84.32}\,{\scriptstyle \pm 0.09}\) & \(\underline{82.38}\,{\scriptstyle \pm 0.08}\) & \(\underline{81.57}\) \\
 & \textsc{TabDiff} & \(81.15\,{\scriptstyle \pm 0.02}\) & \(76.56\,{\scriptstyle \pm 0.05}\) & \(83.65\,{\scriptstyle \pm 0.11}\) & \(\mathbf{83.43}\,{\scriptstyle \pm 0.15}\) & \(72.80\,{\scriptstyle \pm 0.09}\) & \(84.29\,{\scriptstyle \pm 0.28}\) & \(79.43\,{\scriptstyle \pm 0.05}\) & \(80.19\) \\
\midrule[0.2pt]
\multirow{3}{*}[0pt]{\rotatebox[origin=c]{90}{\fontsize{5.5}{6.6}\selectfont Cross-dataset}} & \textsc{TabEBM} & \(69.55\,{\scriptstyle \pm 0.52}\) & \(67.99\,{\scriptstyle \pm 0.47}\) & \(69.70\,{\scriptstyle \pm 0.16}\) & \(71.96\,{\scriptstyle \pm 0.45}\) & \(60.12\,{\scriptstyle \pm 0.25}\) & \(78.97\,{\scriptstyle \pm 0.15}\) & \(62.98\,{\scriptstyle \pm 3.99}\) & \(68.75\) \\
 & \textsc{CTSyn} & \(73.26\,{\scriptstyle \pm 0.41}\) & \(71.72\,{\scriptstyle \pm 0.44}\) & \(78.07\,{\scriptstyle \pm 0.07}\) & \(69.54\,{\scriptstyle \pm 1.18}\) & \(62.81\,{\scriptstyle \pm 0.04}\) & \(82.86\,{\scriptstyle \pm 0.05}\) & \(79.60\,{\scriptstyle \pm 0.12}\) & \(73.98\) \\
 & \textsc{CDMD (Ours)} & \(\mathbf{83.77}\,{\scriptstyle \pm 0.19}\) & \(\mathbf{79.12}\,{\scriptstyle \pm 0.24}\) & \(\mathbf{83.99}\,{\scriptstyle \pm 0.19}\) & \(83.12\,{\scriptstyle \pm 0.04}\) & \(\mathbf{75.84}\,{\scriptstyle \pm 0.04}\) & \(84.08\,{\scriptstyle \pm 0.22}\) & \(\mathbf{82.45}\,{\scriptstyle \pm 0.05}\) & \(\mathbf{81.77}\) \\
\bottomrule
\end{tabular}
}%
\vspace{-4pt}
\end{table}

\custompar{Overview}
We evaluate \ours{} in two complementary settings.
First, we study \textit{in-domain cross-dataset generation} (\Secref{sec:exp:cross_dataset_generation}): can a single model trained jointly on multiple tabular datasets generate high-quality synthetic data for each dataset?
Second, we study \textit{generation in constrained settings} (\Secref{sec:exp:constrained_generation}): can cross-dataset pre-training improve generation on previously unseen datasets when either the amount of target training data or the number of training epochs is limited?
We begin by describing the experimental setup shared by both evaluation settings with additional implementation details in Appendix~\ref{app:implementation_details}.

\custompar{Evaluation datasets}
We use seven real-world tabular datasets widely adopted in tabular generation works \citep{zhang2024mixed,shi2025tabdiff} as our target datasets for evaluation: \textit{Adult}, \textit{Default}, \textit{Magic}, \textit{Shoppers}, \textit{Diabetes}, \textit{Beijing}, and \textit{News}. They span diverse domains, contain different combinations of numerical and categorical features, and each comes with an inherent machine learning task: binary classification, multiclass classification, or regression.
We split each dataset into two equally sized subsets: a training set for fitting and developing the generative models, and a held-out test set for evaluating synthetic data quality.
Detailed dataset information is in Appendix~\ref{app:evaluation_datasets}.

\custompar{Evaluation metrics}
Following \citet{zhang2024mixed, shi2025tabdiff}, we assess synthetic data quality using seven metrics across three evaluation axes \citep{sdmetrics,alaa2022faithful}.

\textbf{(1)} \textit{Distributional Fidelity} ($F$): \textit{Shape} and \textit{Trend} compare real and synthetic column marginals and pairwise column correlations, respectively, whereas \textit{$\alpha$-Precision} and \textit{$\beta$-Recall} assess sample-level resemblance to the real distribution and its coverage, respectively.
We use the held-out test set as the reference data for these metrics and define the fidelity score, $F$, as their average.
\textbf{(2)} \textit{Predictive Utility} ($U$): \textit{Machine Learning Efficacy (MLE)} measures how well synthetic data preserves information useful for the dataset’s prediction task. Following \citet{zhang2024mixed}, we train an XGBoost classifier or regressor on synthetic data and evaluate it on the held-out test data. The utility score, $U$, is based on AUC and normalized RMSE for classification and regression datasets, respectively.
\textbf{(3)} \textit{Privacy proxies} ($P$): We use a \textit{distance-to-closest-record (DCR)} score to assess whether synthetic rows are disproportionately close to the training set relative to the equally sized test set and \textit{Authenticity} to assess potential copying through each synthetic row’s proximity to its nearest training record relative to local training-data distances. These metrics provide empirical diagnostics of potential memorization, and their mean defines the privacy proxy score $P$.
We express all individual metrics on a $0$ to $100$ scale, with higher values indicating better performance.
We provide detailed metric computation and normalization in Appendix~\ref{app:evaluation_metrics}.
Due to space constraints, we report aggregated scores for our experiments in the main text and provide detailed results in Appendix~\ref{app:detailed_results}.
We aggregate the Fidelity and Utility scores into a \textit{Generation Quality} score ($Q$) and the three axes' scores into an \textit{Overall Evaluation Score} ($S$). Averaging first within each axis and then across axes gives each evaluation dimension equal weight, independently of the number of metrics used to measure it. Formally,
\begin{equation*}
    Q=(F+U)/2
    \quad\text{(\textit{Generation Quality})},
    \qquad
    S=(F+U+P)/3
    \quad\text{(\textit{Overall Evaluation Score})}.
\end{equation*}

\custompar{Baselines}
We compare \ours{} against nine tabular generation methods, categorized as single-dataset and cross-dataset methods.
\textbf{(1)} \textit{Single-dataset methods}. We select representative methods from different model families: (i) \textit{interpolation-based}, \textsc{SMOTE} \citep{chawla2002smote}; (ii) \textit{VAE-based}, \textsc{TVAE} \citep{xu2019modeling}; (iii) \textit{GAN-based}, \textsc{CTGAN} \citep{xu2019modeling}; (iv) \textit{language model-based}, \textsc{GReaT} \citep{borisov2022language}; and (v) \textit{diffusion-based}, \textsc{TabDDPM} \citep{kotelnikov2023tabddpm}, \textsc{TabSyn} \citep{zhang2024mixed}, and \textsc{TabDiff} \citep{shi2025tabdiff}.
\textbf{(2)} \textit{Cross-dataset methods}. We evaluate \textsc{TabEBM} \citep{margeloiu2024tabebm}, an energy-based method built on a pretrained in-context classifier, \textsc{TabPFN} \citep{hollmann2022tabpfn}, and \textsc{CTSyn} \citep{lin2025ctsyn}, a schema-conditional cross-dataset generator.
At the time of writing, we could not locate a publicly available implementation of \textsc{LaTable} \citep{van2024latable}, so we exclude it from our evaluations.

\subsection{In-domain cross-dataset generation}
\label{sec:exp:cross_dataset_generation}

\begin{figure}[t]
    \centering
    \includegraphics[width=\linewidth]{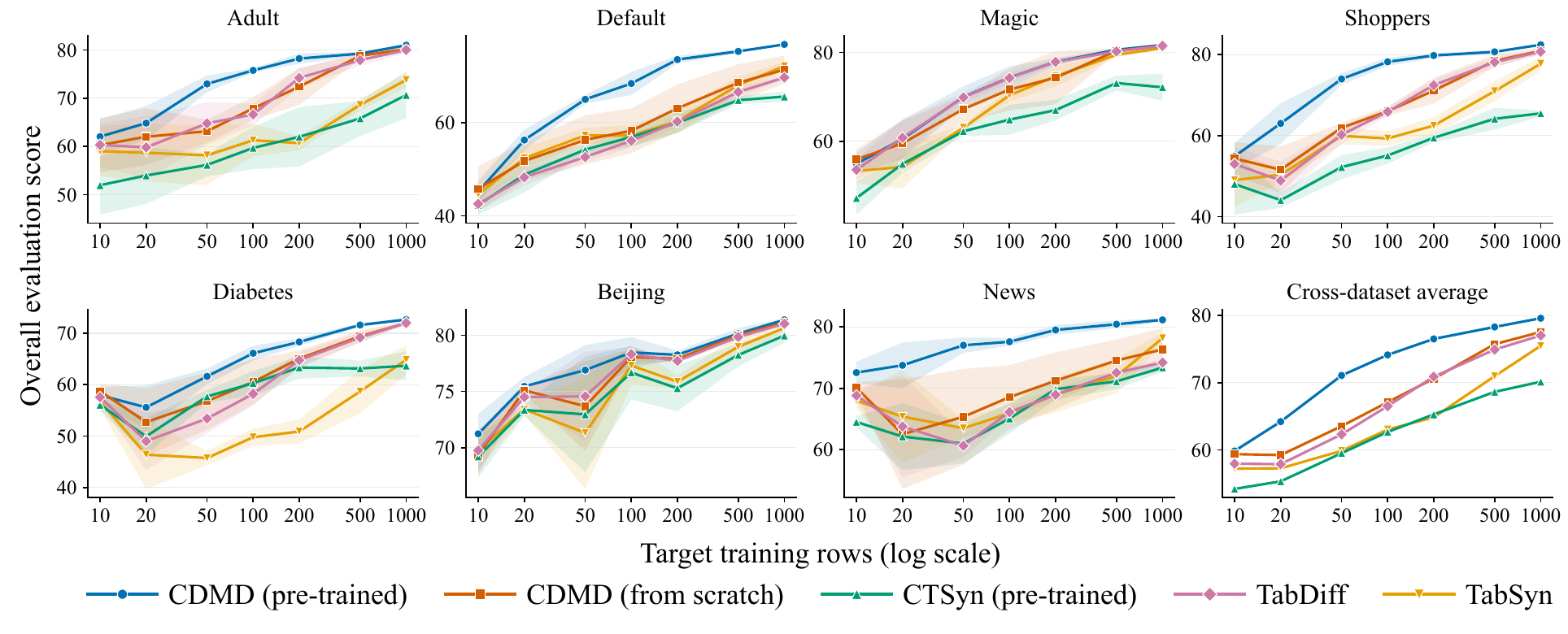}
    \caption{Sample-constrained generation results: Overall Evaluation Score ($S$) vs. the number of target training rows. Lines and shaded regions show the mean $\scriptstyle\pm$ standard deviation over five runs using independently-sampled target training subsets. The final panel shows the mean across datasets.}
    \label{fig:few-shot-performance}
    \vspace{-9pt}
\end{figure}

This experiment tests whether a single \ours{} checkpoint can generate high-quality synthetic data across multiple datasets.
To this end, we train \ours{} jointly on the training splits of the seven evaluation datasets and use the resulting checkpoint to generate synthetic data for each dataset conditioned on its schema.
We follow the same training protocol for \textsc{CTSyn}, while \textsc{TabEBM} uses the corresponding training data as context, and the single-dataset baselines are applied separately to each dataset's training split.
We then evaluate all methods using the same evaluation protocol.

Table~\ref{tab:cross-table-results} shows that \ours{} achieves the highest average generation quality score, 85.09, followed by \textsc{TabSyn} (84.27) and \textsc{TabDiff} (84.22), and obtains the best generation-quality score on six out of seven datasets. It also achieves the highest average overall score, 81.77, followed by \textsc{TabSyn} at 81.57, and substantially outperforms jointly trained \textsc{CTSyn} on both scores.
The detailed results in Appendix~\ref{app:detailed_cross_dataset} clarify this advantage: \ours{} has higher average distributional fidelity than \textsc{TabSyn} and \textsc{TabDiff}, while matching TabDiff’s average predictive utility.
Notably, \ours{} achieves this performance while using a fraction of the total trainable parameters compared to the best single-dataset baselines: \textsc{TabSyn} and \textsc{TabDiff} both have 74.8M total trainable parameters across the seven datasets, while \ours{} has 10.7M parameters (excluding the frozen text embedding model, which is used once to get schema embeddings and is not needed during training nor sampling).
These results demonstrate that a single \ours{} checkpoint can match or exceed strong, separately trained generators while supporting heterogeneous schemas.


\subsection{Generation in constrained settings}
\label{sec:exp:constrained_generation}

\begin{figure}[t]
    \centering
    \includegraphics[width=\linewidth]{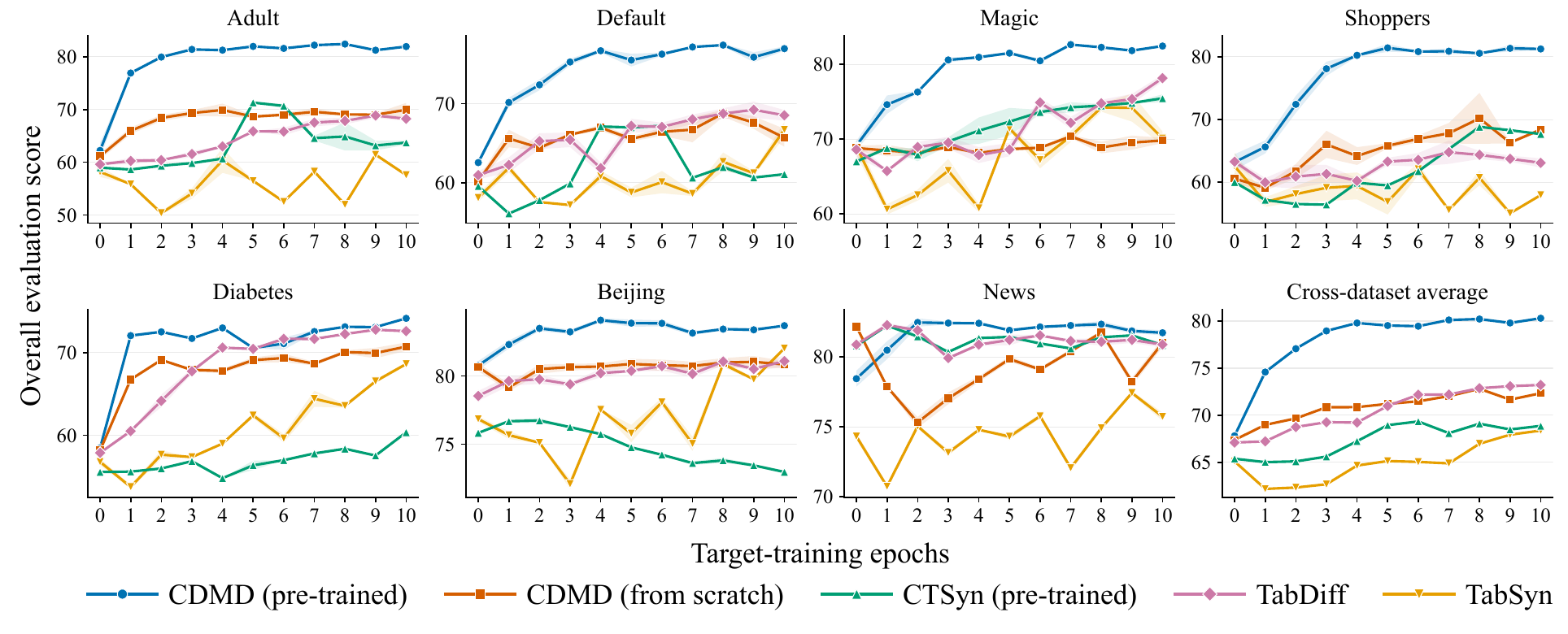}
    \caption{Epoch-constrained generation results: Overall Evaluation Score ($S$) vs. training epochs. Lines and shaded regions indicate the mean $\scriptstyle\pm$ standard deviation over three runs.}
    \label{fig:few-epoch-performance}
    \vspace{-10pt}
\end{figure}

This experiment tests whether cross-dataset transfer can improve generation quality when training data or epochs are limited. Following \citet{van2024latable} and \citet{lin2025ctsyn}, we adopt a pre-training-and-fine-tuning setup. We pre-train \ours{} (a different instance from the previous experiment) on a corpus of 337 tabular datasets, filtered by \citet{yan2024making}. We exclude the evaluation datasets from this corpus through automated checks followed by manual verification. Further corpus details are in Appendix~\ref{app:pre_training_datasets}.
Starting from the same pretrained checkpoint for each run, we fine-tune \ours{} separately on each evaluation dataset, using either a subset of the training set or the full training set for a limited number of epochs.
We compare against \textsc{CTSyn} pretrained on the same corpus, \textsc{TabSyn} and \textsc{TabDiff} as the strongest single-dataset baselines from the previous experiment, and \ours{} trained from scratch. 
In the data-constrained setting, all methods use the same target training rows, and in the epoch-constrained setting, they use the same number of epochs.

\custompar{Adaptation with limited target data}
For each evaluation dataset and sample budget $N\in\{10,20,50,100,200,500,1000\}$, we draw five independent random subsets of $N$ rows from the training split to account for variability in target-data selection. Within each draw, all methods, including numerical pre-processing, are fitted on exactly the same rows. For each run, we generate a synthetic dataset matching the size of the held-out test set and compute the evaluation scores (for DCR scores, we subsample the held-out set to $N$ and report the average over 20 seeds). 
Keeping the synthetic sample count fixed across budgets controls for its effect on the metrics and maintains consistency with the previous experiment.
\Figref{fig:few-shot-performance} shows that pre-trained \ours{} achieves the highest cross-dataset-average overall score at every sample budget, with the clearest advantage at tens to a few hundred target examples.
Improvements are particularly evident on Adult, Default, Shoppers, and News. 
As more target rows become available, the average gap generally narrows as models trained from scratch improve. The generation-quality results in Appendix~\ref{app:few_shot_generation} show that the benefit also appears when privacy proxies are excluded from the aggregate. These results demonstrate that cross-dataset pretraining supports effective generation from limited target data.

\custompar{Adaptation with limited training epochs}
We next train or adapt each method on the full training split of each target dataset and evaluate generation quality over the first ten epochs, with epoch 0 denoting no adaptation/training. An epoch corresponds to one complete pass through the training rows (by both components for two-stage models), regardless of batch size. This setting assesses how pretraining affects generation quality during the early stages of training on a new dataset.
As shown in \Figref{fig:few-epoch-performance}, pretrained \ours{} improves rapidly during the initial epochs and approaches its observed plateau within a few epochs. It attains the highest cross-dataset average overall score after one epoch and retains this advantage through epoch ten, with clear improvements on Adult, Default, Magic, and Shoppers. The advantage over \ours{} trained from scratch supports the benefit of pretrained adaptation within a limited number of target-training epochs. Appendix~\ref{app:few_epoch_generation} shows the corresponding generation-quality curves, which support the same conclusion without the privacy proxies. Together with the limited-data experiment, these results establish the value of pretrained \ours{} under both constraints: scarce target observations and limited adaptation epochs.

\section{Conclusion}
In this work, we introduced \ours{}, a mixed-type diffusion model that can be trained end-to-end across heterogeneous tabular datasets.
By combining numerical and categorical diffusion with a shared schema-conditioned denoiser, \ours{} accommodates varying feature sets and categorical vocabularies directly in the mixed-type feature space.
Experiments on seven real-world datasets show that a single jointly trained checkpoint matches or outperforms strong separately trained generators on average. Pretraining on 337 datasets further improves generation on unseen datasets under limited target data and adaptation epochs.

\custompar{Limitations and future work}
Our current adaptation strategy relies on fine-tuning, requiring additional optimization for each target dataset. Future work could extend \textsc{CDMD} to in-context generation, enabling adaptation by conditioning on example rows without updating model parameters. Our experiments are also limited in scale, leaving performance at larger scales an open question. Nevertheless, the shared schema-conditional architecture provides a starting point for scaling tabular generative models to larger and more diverse corpora.


\newpage

\subsubsection*{Acknowledgments}
This project was funded by SAP SE.
It is also supported by the DAAD programme Konrad Zuse Schools of Excellence in Artificial Intelligence, sponsored by the Federal Ministry of Education and Research.
We thank Mojtaba Nayyeri, Johannes Hoffart, and Filippo Guerranti for their valuable feedback.

\bibliography{iclr2027_conference}

@article{shi2025comprehensive,
  title={A comprehensive survey of synthetic tabular data generation},
  author={Shi, Ruxue and Wang, Yili and Du, Mengnan and Shen, Xu and Chang, Yi and Wang, Xin},
  journal={arXiv preprint arXiv:2504.16506},
  year={2025}
}

@article{xu2019modeling,
  title={Modeling tabular data using conditional gan},
  author={Xu, Lei and Skoularidou, Maria and Cuesta-Infante, Alfredo and Veeramachaneni, Kalyan},
  journal={Advances in neural information processing systems},
  volume={32},
  year={2019}
}

@article{borisov2022language,
  title={Language models are realistic tabular data generators},
  author={Borisov, Vadim and Se{\ss}ler, Kathrin and Leemann, Tobias and Pawelczyk, Martin and Kasneci, Gjergji},
  journal={arXiv preprint arXiv:2210.06280},
  year={2022}
}

@inproceedings{kotelnikov2023tabddpm,
  title={Tabddpm: Modelling tabular data with diffusion models},
  author={Kotelnikov, Akim and Baranchuk, Dmitry and Rubachev, Ivan and Babenko, Artem},
  booktitle={International conference on machine learning},
  pages={17564--17579},
  year={2023},
  organization={PMLR}
}

@article{van2024tabular,
  title={Why tabular foundation models should be a research priority},
  author={Van Breugel, Boris and Van Der Schaar, Mihaela},
  journal={arXiv preprint arXiv:2405.01147},
  year={2024}
}

@article{hollmann2025accurate,
  title={Accurate predictions on small data with a tabular foundation model},
  author={Hollmann, Noah and M{\"u}ller, Samuel and Purucker, Lennart and Krishnakumar, Arjun and K{\"o}rfer, Max and Hoo, Shi Bin and Schirrmeister, Robin Tibor and Hutter, Frank},
  journal={Nature},
  volume={637},
  number={8045},
  pages={319--326},
  year={2025},
  publisher={Nature Publishing Group UK London}
}

@article{qu2025tabicl,
  title={Tabicl: A tabular foundation model for in-context learning on large data},
  author={Qu, Jingang and Holzm\"uller, David and Varoquaux, Ga\"el and Morvan, Marine Le},
  journal={arXiv preprint arXiv:2502.05564},
  year={2025}
}

@article{margeloiu2024tabebm,
  title={Tabebm: A tabular data augmentation method with distinct class-specific energy-based models},
  author={Margeloiu, Andrei and Jiang, Xiangjian and Simidjievski, Nikola and Jamnik, Mateja},
  journal={Advances in Neural Information Processing Systems},
  volume={37},
  pages={72094--72144},
  year={2024}
}

@article{van2024latable,
  title={LaTable: towards large tabular models},
  author={van Breugel, Boris and Crabb{\'e}, Jonathan and Davis, Rob and van der Schaar, Mihaela},
  journal={arXiv preprint arXiv:2406.17673},
  year={2024}
}

@inproceedings{lin2025ctsyn,
  title={CTSyn: A foundation model for cross tabular data generation},
  author={Lin, Xiaofeng and Xu, Chenheng and Yang, Matthew and Cheng, Guang},
  booktitle={International Conference on Learning Representations},
  volume={2025},
  pages={47938--47959},
  year={2025}
}

@article{ma2024tabpfgen,
  title={TabPFGen--Tabular Data Generation with TabPFN},
  author={Ma, Junwei and Dankar, Apoorv and Stein, George and Yu, Guangwei and Caterini, Anthony},
  journal={arXiv preprint arXiv:2406.05216},
  year={2024}
}

@article{song2020score,
  title={Score-based generative modeling through stochastic differential equations},
  author={Song, Yang and Sohl-Dickstein, Jascha and Kingma, Diederik P and Kumar, Abhishek and Ermon, Stefano and Poole, Ben},
  journal={arXiv preprint arXiv:2011.13456},
  year={2020}
}

@article{sahoo2024simple,
  title={Simple and effective masked diffusion language models},
  author={Sahoo, Subham S and Arriola, Marianne and Schiff, Yair and Gokaslan, Aaron and Marroquin, Edgar and Chiu, Justin T and Rush, Alexander and Kuleshov, Volodymyr},
  journal={Advances in Neural Information Processing Systems},
  volume={37},
  pages={130136--130184},
  year={2024}
}

@article{hollmann2022tabpfn,
  title={Tabpfn: A transformer that solves small tabular classification problems in a second},
  author={Hollmann, Noah and M{\"u}ller, Samuel and Eggensperger, Katharina and Hutter, Frank},
  journal={arXiv preprint arXiv:2207.01848},
  year={2022}
}

@article{jiang2026tabular,
  title={Tabular Foundation Model for Generative Modelling},
  author={Jiang, Xiangjian and Liu, Mingxuan and Simidjievski, Nikola and Klein, Tassilo and Jamnik, Mateja},
  journal={arXiv preprint arXiv:2605.09424},
  year={2026}
}

@inproceedings{sohl2015deep,
  title={Deep unsupervised learning using nonequilibrium thermodynamics},
  author={Sohl-Dickstein, Jascha and Weiss, Eric and Maheswaranathan, Niru and Ganguli, Surya},
  booktitle={International conference on machine learning},
  pages={2256--2265},
  year={2015},
  organization={pmlr}
}

@article{austin2021structured,
  title={Structured denoising diffusion models in discrete state-spaces},
  author={Austin, Jacob and Johnson, Daniel D and Ho, Jonathan and Tarlow, Daniel and Van Den Berg, Rianne},
  journal={Advances in neural information processing systems},
  volume={34},
  pages={17981--17993},
  year={2021}
}

@inproceedings{zhang2024mixed,
  title={Mixed-type tabular data synthesis with score-based diffusion in latent space},
  author={Zhang, Hengrui and Zhang, Jiani and Shen, Zhengyuan and Srinivasan, Balasubramaniam and Qin, Xiao and Faloutsos, Christos and Rangwala, Huzefa and Karypis, George},
  booktitle={International Conference on Learning Representations},
  volume={2024},
  pages={52829--52857},
  year={2024}
}

@inproceedings{shi2025tabdiff,
  title={Tabdiff: a mixed-type diffusion model for tabular data generation},
  author={Shi, Juntong and Xu, Minkai and Hua, Harper and Zhang, Hengrui and Ermon, Stefano and Leskovec, Jure},
  booktitle={International Conference on Learning Representations},
  volume={2025},
  pages={37353--37375},
  year={2025}
}

@article{vaswani2017attention,
  title={Attention is all you need},
  author={Vaswani, Ashish and Shazeer, Noam and Parmar, Niki and Uszkoreit, Jakob and Jones, Llion and Gomez, Aidan N and Kaiser, {\L}ukasz and Polosukhin, Illia},
  journal={Advances in neural information processing systems},
  volume={30},
  year={2017}
}

@article{li2023towards,
  title={Towards general text embeddings with multi-stage contrastive learning},
  author={Li, Zehan and Zhang, Xin and Zhang, Yanzhao and Long, Dingkun and Xie, Pengjun and Zhang, Meishan},
  journal={arXiv preprint arXiv:2308.03281},
  year={2023}
}

@article{chawla2002smote,
  title={SMOTE: synthetic minority over-sampling technique},
  author={Chawla, Nitesh V and Bowyer, Kevin W and Hall, Lawrence O and Kegelmeyer, W Philip},
  journal={Journal of artificial intelligence research},
  volume={16},
  pages={321--357},
  year={2002}
}

@manual{
    sdmetrics,
    title = {Synthetic Data Metrics},
    author = {DataCebo, Inc.},
    organization = {DataCebo, Inc.},
    year = {2026},
    month = {02},
    note = {Version 0.27.1},
    url = {https://docs.sdv.dev/sdmetrics/},
}

@inproceedings{alaa2022faithful,
  title={How faithful is your synthetic data? sample-level metrics for evaluating and auditing generative models},
  author={Alaa, Ahmed and Van Breugel, Boris and Saveliev, Evgeny S and Van Der Schaar, Mihaela},
  booktitle={International conference on machine learning},
  pages={290--306},
  year={2022},
  organization={PMLR}
}

@article{qian2023synthcity,
  title={Synthcity: facilitating innovative use cases of synthetic data in different data modalities},
  author={Qian, Zhaozhi and Cebere, Bogdan-Constantin and van der Schaar, Mihaela},
  journal={arXiv preprint arXiv:2301.07573},
  year={2023}
}

@article{karras2022elucidating,
  title={Elucidating the design space of diffusion-based generative models},
  author={Karras, Tero and Aittala, Miika and Aila, Timo and Laine, Samuli},
  journal={Advances in neural information processing systems},
  volume={35},
  pages={26565--26577},
  year={2022}
}

@article{han2026breaking,
  title={Breaking the Quality-Privacy Tradeoff in Tabular Data Generation via In-Context Learning},
  author={Han, Xinyan and Lu, Yan and Lin, Xiaoyu and Jiang, Yuanyuan and Wang, Yuanrui and Li, Xuanyue and Zou, Wenchao and Zhang, Xingxuan},
  journal={arXiv preprint arXiv:2605.04911},
  year={2026}
}

@article{van2023beyond,
  title={Beyond privacy: Navigating the opportunities and challenges of synthetic data},
  author={Van Breugel, Boris and Van der Schaar, Mihaela},
  journal={arXiv preprint arXiv:2304.03722},
  year={2023}
}

@article{borisov2022deep,
  title={Deep neural networks and tabular data: A survey},
  author={Borisov, Vadim and Leemann, Tobias and Se{\ss}ler, Kathrin and Haug, Johannes and Pawelczyk, Martin and Kasneci, Gjergji},
  journal={IEEE transactions on neural networks and learning systems},
  volume={35},
  number={6},
  pages={7499--7519},
  year={2022},
  publisher={IEEE}
}

@article{stoian2025survey,
  title={A survey on deep learning approaches for tabular data generation: Utility, alignment, fidelity, privacy, diversity, and beyond},
  author={Stoian, Mihaela Catalina and Giunchiglia, Eleonora and Lukasiewicz, Thomas},
  journal={arXiv preprint arXiv:2503.05954},
  year={2025}
}

@article{cai2026tabdlm,
  title={TabDLM: Free-Form Tabular Data Generation via Joint Numerical-Language Diffusion},
  author={Cai, Donghong and Feng, Jiarui and Wang, Yanbo and Zheng, Da and Chen, Yixin and Zhang, Muhan},
  journal={arXiv preprint arXiv:2602.22586},
  year={2026}
}

@article{zhang2025limix,
  title={Limix: Unleashing structured-data modeling capability for generalist intelligence},
  author={Zhang, Xingxuan and Ren, Gang and Yu, Han and Yuan, Hao and Wang, Hui and Li, Jiansheng and Wu, Jiayun and Mo, Lang and Mao, Li and Hao, Mingchao and others},
  journal={arXiv preprint arXiv:2509.03505},
  year={2025}
}

@inproceedings{yan2024making,
  title={Making pre-trained language models great on tabular prediction},
  author={Yan, Jiahuan and Zheng, Bo and Xu, Hongxia and Zhu, Yiheng and Chen, Danny and Sun, Jimeng and Wu, Jian and Chen, Jintai},
  booktitle={International Conference on Learning Representations},
  volume={2024},
  pages={23570--23593},
  year={2024}
}

@inproceedings{ye2024towards,
  title={Towards cross-table masked pretraining for web data mining},
  author={Ye, Chao and Lu, Guoshan and Wang, Haobo and Li, Liyao and Wu, Sai and Chen, Gang and Zhao, Junbo},
  booktitle={Proceedings of the ACM Web Conference 2024},
  pages={4449--4459},
  year={2024}
}
\bibliographystyle{iclr2027_conference}

\newpage

\appendix
\section{Related work}
\label{sec:related_work}

\custompar{Single-dataset tabular generative models} 
Tabular data generation must accommodate substantial heterogeneity across columns within a single dataset: features differ in data type, numerical scale, marginal distribution, and categorical vocabulary, while their dependencies determine the joint row distribution. Neural approaches such as \textsc{TVAE} and \textsc{CTGAN}~\citep{xu2019modeling} address mixed numerical and categorical data through dedicated representations; \textsc{CTGAN} additionally uses mode-specific normalization and conditional sampling to handle multimodal numerical distributions and imbalanced categories. Diffusion models offer several approaches to this heterogeneity. \textsc{TabDDPM}~\citep{kotelnikov2023tabddpm} combines Gaussian diffusion for numerical features with multinomial diffusion for categorical features, using a joint denoiser to capture dependencies. \textsc{TabSyn}~\citep{zhang2024mixed} instead learns a Transformer-based variational autoencoder and applies score-based diffusion in its continuous latent space. Most closely related to our mixed-type formulation, \textsc{TabDiff}~\citep{shi2025tabdiff} jointly models numerical and categorical features using Gaussian and masked diffusion models, with learnable, feature-specific schedules that accommodate differences in column distributions. \textsc{TabDLM}~\citep{cai2026tabdlm} further combines numerical diffusion with masked diffusion over categorical and free-form text tokens using a pretrained diffusion language model. These methods are fitted to individual datasets in their reported generative experiments. \ours{} addresses the additional challenge of sharing a mixed-type generator across datasets with different schemas and categorical vocabularies.

\custompar{Cross-dataset tabular generative models} Cross-dataset generators seek to transfer generative knowledge across tables despite differences in feature semantics, types, and dimensionality. \textsc{LaTable}~\citep{van2024latable} uses a shared Transformer conditioned on textual dataset and column information. It applies Gaussian diffusion to numerical values and language-model embeddings of categorical values, recovering categorical probabilities through similarity scores against each feature’s vocabulary. \textsc{CTSyn}~\citep{lin2025ctsyn} learns an autoencoder that maps heterogeneous rows to a common latent representation, together with a schema-guided decoder that reconstructs numerical and categorical values. A schema-conditioned latent diffusion model generates these representations, supporting pretraining across tables and subsequent adaptation. \textsc{DiffICL}~\citep{han2026breaking} adopts an in-context formulation: a frozen \textsc{LimiX}~\citep{zhang2025limix} encoder represents target-table examples, and a pretrained latent diffusion model generates new representations conditioned on context rows. Its reconstruction stage still requires dataset-specific training, fitting a separate lightweight decoder for each feature using target-table data. Thus, although the diffusion model is shared, the complete generation pipeline includes newly trained target-specific components. These approaches perform diffusion in continuous representations. \ours{} defines diffusion directly over mixed numerical and categorical states, with schema-conditioned modules shared across datasets and categorical predictions restricted to each feature’s vocabulary.

\custompar{Tabular foundation models and their generative adaptations} Predictive tabular foundation models, including \textsc{TabPFN}~\citep{hollmann2022tabpfn,hollmann2025accurate} and \textsc{TabICL}~\citep{qu2025tabicl}, learn transferable prediction procedures through pretraining across (synthetic) tasks. They use labeled examples from a target dataset as context to predict unseen targets without updating model parameters. Their standard objective is conditional target prediction, while synthetic row generation requires a model of the joint distribution over features. \citet{hollmann2025accurate} show that \textsc{TabPFN} can nevertheless serve as an autoregressive generator by treating successive columns as prediction targets and sampling each feature conditional on previously generated features and real-data context. This construction reuses pretrained conditional predictors and introduces feature ordering into the generation process.
Other adaptations construct explicit generative models from pretrained predictors. \textsc{TabPFGen}~\citep{ma2024tabpfgen} derives a class-conditional energy from frozen \textsc{TabPFN} outputs and generates samples using stochastic gradient Langevin dynamics. \textsc{TabEBM}~\citep{margeloiu2024tabebm} constructs distinct class-specific energy models through surrogate binary classification tasks evaluated by a pretrained in-context classifier. \textsc{TabFORGE}~\citep{jiang2026tabular} uses frozen representations from a pretrained \textsc{TabPFN} encoder, pretrains a latent diffusion Transformer, and subsequently trains a decoder on denoised representations. Its reconstruction interface contains table-specific detokenizers with feature-specific output heads; adaptation initializes new heads and fine-tunes the pretrained diffusion and decoder networks. \ours{} learns its generative model directly in the mixed-type feature space, using shared schema-conditioned input and output modules without relying on a pretrained tabular predictor.

\section{Method Details}
\label{app:method_details}

\begin{figure}[t]
    \centering
    \includegraphics[width=\linewidth]{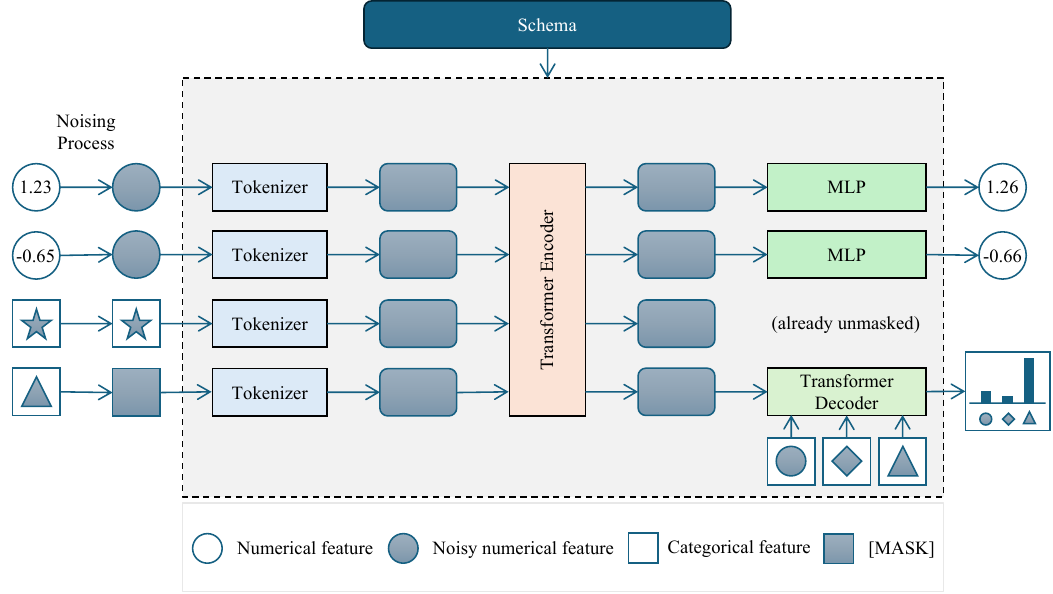}
    \caption{Overview of \ours{}. Numerical features are corrupted with Gaussian noise, while categorical features are randomly masked. The tokenizer combines each noisy feature with schema information, and a shared transformer encoder captures dependencies across columns. A shared MLP predicts clean numerical values, while a shared transformer decoder uses category embeddings and the contextualized feature token to produce probabilities over each categorical column’s valid vocabulary. These predictions guide the mixed-type reverse diffusion process. The schema determines the number of feature tokens and categorical outputs, while parameters are shared across datasets and columns of the same type, allowing a single model to handle heterogeneous tables.}
    \label{fig:overview}
\end{figure}



\begin{algorithm}[H]
\caption{Cross-dataset training of \ours{}}\label{alg:training}

\begin{algorithmic}[1]

\Require $L$ datasets $\{ \gT^{(\ell)} \}_{\ell=1}^L = \{ (\gS^{(\ell)}, \mX^{(\ell)}) \}_{\ell=1}^L$, denoising network $\vmu_\vtheta$
\Repeat
\State Sample $\ell\sim p_\mathrm{dataset}(\{ 1,2,\dots,L \})$ \Comment{Sample dataset index}
\State Sample $\vx_0 \sim \gU(\mX^{(\ell)})$ \Comment{Sample clean row}
\State Sample $t\sim \gU([0,1])$ \Comment{Sample diffusion time}
\LineComment{Forward noising process:}
\State Initialize $\vx_t$ with same shape as $\vx_0$
\State $\vx_t^\mathrm{num} \gets \vx_0^\mathrm{num} + \sigma_t \bm{\eps}$ with $\bm{\eps} \sim \gN(\vzero, \mI)$ \Comment{(\Eqref{eq:gaussian_forward_diffusion})}
\State $x_{t,j} \sim q_\mathrm{cat}(x_{t,j}\mid x_{0,j}) \quad \forall j\in \gJ_\mathrm{cat}^{(\ell)}$ \Comment{Mask with probability $1-\alpha_t$ (~\Eqref{eq:masked_forward_diffusion})} 
\LineComment{Network prediction:}
\State $\hat\vx \gets \vmu_\vtheta(\vx_t,t,\gS^{(\ell)})$ 
\State $\gL_\mathrm{num} \gets \frac{\lambda_t}{|\gJ_\mathrm{num}^{(\ell)}|} \left\|\hat\vx^\mathrm{num} - \vx_0^\mathrm{num} \right\|_2^2$
\State $\gL_\mathrm{cat} \gets \frac{\alpha_t'}{|\gJ_\mathrm{cat}^{(\ell)}|(1-\alpha_t)} \sum_{j\in\gJ_\mathrm{cat}^{(\ell)},x_{t,j}=m} \log \left( \hat\vx_j[x_{0,j}] \right)$
\State Take gradient descent step on $\nabla_\vtheta (\gL_\mathrm{num} + \gL_\mathrm{cat})$
\Until{converged.}

\end{algorithmic}
\end{algorithm}

\begin{algorithm}
\caption{Schema-conditioned sampling from \ours{}}\label{alg:sampling}
\begin{algorithmic}[1]
\Require Schema $\gS$, trained denoising network $\vmu_\vtheta$, discretization time steps $\{ t_i:i\in\{ 0,1,\dots,N \} \}$ such that $t_0=0<t_1<\dots<t_N=1$
\State $\vx_1^\mathrm{num} \sim \gN(\vzero_{|\gJ_\mathrm{num}|},\sigma_\mathrm{max}^2\mI_{|\gJ_\mathrm{num}|})$
\State $\forall j\in\gJ_\mathrm{cat}\; x_{1,j}\gets m$
\For{$i\in \{ N,N-1,\dots,1 \}$}
\If{$i<N$}
\State $\hat t_i \gets \min(\sigma^{-1}(\sigma_{t_i}+\frac{1}{N} \sigma_{t_i}),1)$ \Comment{Select temporarily increased noise level $\hat t_i$}
\LineComment{Numerical forward perturbation}
\State $\vx_{\hat t_i}^\mathrm{num} \gets \vx_{t_i}^\mathrm{num} + \sqrt{\sigma_{\hat t_i}^2 - \sigma_{t_i}^2} \bm{\epsilon}_i$ with $\bm{\epsilon}_i \sim \gN(\vzero,\mI)$
\LineComment{Categorical forward perturbation}
\State $\forall j \in \gJ_\mathrm{cat}\; x_{\hat t_i,j}\sim q_\mathrm{cat}(x_{\hat t_i,j}\mid x_{t_i,j})$ \Comment{(\Eqref{eq:masked_step_transition})}
\Else
\State $\hat t_i = t_i$ \Comment{Stochastic sampler is disabled at the first sampling iteration}
\EndIf
\LineComment{Model prediction}
\State $\hat \vx_i \gets \vmu_\vtheta(\vx_{\hat t_i},\hat t_i,\gS)$
\LineComment{Numerical backward sampling}
\State $d_i \gets \frac{\sigma_{\hat t_i}'}{\sigma_{\hat t_i}} (\vx_{\hat t_i}^\mathrm{num} - \hat \vx_i^\mathrm{num})$ \Comment{Evaluate $\frac{d\vx^\mathrm{num}}{dt}$ at $\hat t_i$}
\State $\vx_{t_{i-1}}^\mathrm{num}\gets \vx_{\hat t_i}^\mathrm{num} + (t_{i-1}-\hat t_i)d_i$ \Comment{Take Euler step from $\hat t_i$ to $t_{i-1}$}
\LineComment{Categorical backward sampling} 
\State $\forall j\in\gJ_\mathrm{cat}\; x_{t_{i-1},j} \sim p_\vtheta(x_{t_{i-1},j}\mid x_{\hat t_i,j}, \hat \vx_{i,j})$ \Comment{(\Eqref{eq:parametrized_masked_reverse_process_row})}
\EndFor
\State \Return $\vx_{t_0}$

\end{algorithmic}
\end{algorithm}

\subsection{Diffusion model details}
\label{app:diffusion_details}

\paragraph{Diffusion schedules.}

The Gaussian diffusion schedule $\sigma_t\colon[0,1]\to\R_+$ is an increasing function of $t$. Following \citet{karras2022elucidating} and \citet{shi2025tabdiff}, we adopt the following schedule
\begin{equation}
    \sigma_t = \left( \sigma_\mathrm{min}^{\frac{1}{\rho}} + t \left( \sigma_\mathrm{max}^{\frac{1}{\rho}} - \sigma_\mathrm{min}^{\frac{1}{\rho}} \right) \right)^\rho,
\end{equation}
and we use the default values in \citet{karras2022elucidating} throughout all our experiments: $\sigma_\mathrm{min}=0.002$, $\sigma_\mathrm{max}=80$ and $\rho=7$.

The masking diffusion schedule $\alpha_t:[0,1]\to[0,1]$ is a strictly decreasing function of $t$. We use a simple linear schedule \citep{austin2021structured,sahoo2024simple}
\begin{equation}
    \alpha_t = 1-t.
\end{equation}



\paragraph{Numerical loss weighting.}
The numerical loss weighting $\lambda_t$ is defined as in \citet{karras2022elucidating}
\begin{equation}
    \lambda_t = \frac{\sigma_t^2+\sigma_\mathrm{data}^2}{(\sigma_t\sigma_\mathrm{data})^2}.
\end{equation}

\paragraph{Additional masked diffusion details.}
The step transition induced by the masking forward process from $t'<t$ to $t$ can be written as \citep{sahoo2024simple}
\begin{equation}
\label{eq:masked_step_transition}
    q_\mathrm{cat}(x_t\mid x_{t'}) =  \frac{\alpha_t}{\alpha_{t'}} \1_{x_t=x_{t'}} + \left(1-\frac{\alpha_t}{\alpha_{t'}}\right) \1_{x_t=m}
\end{equation}

The row-conditioned reverse categorical process for a column $j$ can be written as
\begin{equation}
\label{eq:parametrized_masked_reverse_process_row}
    p_\vtheta(x_{t'}\mid x_t, \hat\vx) = 
    \begin{cases}
        \1_{x_{t'}=x_t}, & \text{ if } x_t\neq m, \\
        \frac{\alpha_{t'}-\alpha_t}{1-\alpha_t} \hat\vx[x_{t'}]\1_{x_{t'}\in\gV} + \frac{1-\alpha_{t'}}{1-\alpha_t}\1_{x_{t'}=m}, & \text{ if } x_t=m.
    \end{cases}
\end{equation}

\paragraph{Training dataset sampling distribution.}
When training on $L$ datasets, we sample dataset $\ell$ with probability 
\begin{equation}
    p_\mathrm{dataset}(\ell) = \frac{{N^{(\ell)}}^{\tau_\mathrm{data}}}{\sum_{\ell'=1}^L {N^{(\ell')}}^{\tau_\mathrm{data}}}
\end{equation}
where $N^{(\ell)}$ is the number of training samples and $\tau_{\mathrm{data}}$ controls the influence of dataset size. In the cross-dataset experiments, we set $\tau_{\mathrm{data}}=0$, assigning equal sampling probability to all datasets. During pre-training, we use $\tau_{\mathrm{data}}=0.5$ to accommodate the substantial variation in dataset sizes while limiting the dominance of larger datasets.

\paragraph{Training and sampling.}
We present pseudocode for training and sampling from \ours{} in Algorithms~\ref{alg:training} and~\ref{alg:sampling}, respectively.
Note that Algorithm~\ref{alg:training} and \Eqref{eq:total_loss} assume that each training dataset contains at least one feature of each type. In our implementation, if a dataset contains no features of a given type, we set the corresponding loss to 0.

\subsection{Architecture details}
\label{app:architecture_details}
We use the same model architecture across all experiments. The text embedding dimension is $d_{\mathrm{emb}}=768$, and the token dimension is $D=384$. The Transformer encoder consists of six layers, each with eight attention heads and a feedforward dimension of 768. All MLPs have two layers, with dimensions matched to their respective inputs and outputs. Following TabDiff \citep{shi2025tabdiff}, we adopt EDM network preconditioning for numerical denoising \citep{karras2022elucidating} using their default hyperparameters. All components share parameters across datasets and across columns of the corresponding type. The model has 10,655,234 trainable parameters, excluding the frozen text embedding model.

\subsection{Desiderata for cross-dataset tabular generative models}
\label{app:desiderata}
Ideally, a cross-dataset tabular generative model $p_\vtheta$ should satisfy the following desiderata:

\begin{mytheorem}{Desiderata for Cross-Dataset Tabular Generative Models}
For an arbitrary table schema $\gS$, $p_\vtheta(\cdot\mid \gS)$ should satisfy the following desiderata:
\begin{enumerate}[leftmargin=*]
    \item \textbf{Schema-Valid Support}:
    Rows not consistent with the schema have zero probability,
        \begin{equation*}
            \forall \vx \notin \gX_\gS,\, p_\vtheta(\vx\mid \gS) = 0.
        \end{equation*}
    \item \textbf{Column Permutation Invariance}:
    For any permutation function $\pi$ over the columns, let $\vx_\pi$ and $\gS_\pi$ denote the permuted row and schema, respectively. Then, the probability of a given row remains invariant to column permutation,
        \begin{equation*}
            \forall \vx \in \gX_\gS,\, p_\vtheta(\vx_\pi|\gS_\pi)=p_\vtheta(\vx\mid \gS).
        \end{equation*}
\end{enumerate}
\end{mytheorem}
\medskip 

\ours{} satisfies both desiderata by construction. Schema-valid support follows from the type-specific reverse processes: numerical values remain in $\R$, while categorical predictions are restricted to each column’s valid vocabulary, with zero mask probability at $t=0$. Shared feature-wise tokenization and detokenization, together with a transformer encoder without positional encodings, make the denoiser equivariant to joint column and schema-entry permutations. Since the initialization and reverse-diffusion updates preserve this symmetry, the resulting distribution is column-permutation invariant.
\section{Experimental Details}
\label{app:experimental_details}

\subsection{Metrics}
\label{app:evaluation_metrics}

\paragraph{Shape.}
Shape measures the statistical similarity between column-marginal distributions of the real and synthetic data. It is based on the Kolmogorov-Smirnov (KS) test for numerical features and the Total Variation (TV) distance for categorical features.
Formally, let $\mX^r$ and $\mX^s$ denote the real and synthetic tables, respectively. The KS statistic corresponding to a numerical column $j$ is defined as 
\begin{equation*}
    \text{KS}(\mX_{:,j}^r, \mX_{:,j}^s) = \sup_u |\hat F_j^r(u) - \hat F_j^s(u)|,
\end{equation*}
where $\hat F_j^r$ and $\hat F_j^s$ are the empirical cumulative distribution functions (CDF) estimated from $\mX_{:,j}^r$ and $\mX_{:,j}^s$, respectively.
For a categorical column $j$, the TV distance is defined as
\begin{equation*}
    \text{TV}(\mX_{:,j}^r, \mX_{:,j}^s) = \frac{1}{2} \sum_{v\in\gV_j} |\hat p_j^r(v)- \hat p_j^s(v)|,
\end{equation*}
where $\hat p_j^r$ and $\hat p_j^s$ are the empirical probability mass functions (PMF) estimated from $\mX_{:,j}^r$ and $\mX_{:,j}^s$, respectively, and $\gV_j$ is the vocabulary set of column $j$.
The final Shape score is computed by aggregating the complements of these tests across all columns as follows
\begin{equation*}
    \text{Shape}_j = 
    \begin{cases}
        1-\text{KS}(\mX_{:,j}^r, \mX_{:,j}^s) & \text{if } \tau_j=\mathrm{num} \\
        1-\text{TV}(\mX_{:,j}^r, \mX_{:,j}^s) & \text{if } \tau_j=\mathrm{cat}
    \end{cases}
    \qquad
    \text{Shape}(\mX^r,\mX^s) = 100\times\frac{1}{d} \sum_{j=1}^d \text{Shape}_j
\end{equation*}

\paragraph{Trend.}
Trend compares column-pair correlations between the real and synthetic data. For two numerical columns $j$ and $k$, we compare their empirical Pearson correlations, $\rho_{jk}^r$ and $\rho_{jk}^s$, estimated from the real and synthetic set, respectively, yielding the distance
\begin{equation*}
    D(\mX_{:,jk}^r, \mX_{:,jk}^s) = \frac{1}{2} |\rho_{jk}^r - \rho_{jk}^s|.
\end{equation*}
For categorical pairs, we compare their normalized contingency tables
\begin{equation*}
    D(\mX_{:,jk}^r, \mX_{:,jk}^s) = \frac{1}{2} \sum_{v\in\gV_j,v'\in\gV_k} |\hat p_{jk}^r(v,v') - \hat p_{jk}^s(v,v')|,
\end{equation*}
where $\hat p_{jk}^r(v,v')$ and $\hat p_{jk}^s(v,v')$ are the proportions of rows containing the category combination $(v,v')$ in the real and synthetic sets, respectively. 
Mixed numerical-categorical pairs use the same contingency computation after discretizing the numerical column.
The Trend score is obtained by normalizing and aggregating these distances across different pairs as follows
\begin{equation*}
    \mathrm{Trend}_{jk} = 1 - D(\mX_{:,jk}^r, \mX_{:,jk}^s), \qquad \mathrm{Trend} = 100\times \frac{1}{|\gP|} \sum_{(j,k)\in\gP} \mathrm{Trend}_{jk},
\end{equation*}
where $\gP$ contains unordered column pairs with real data correlation strength above a pre-defined threshold.
Both metrics are normalized to $[0,100]$, with higher values indicating greater similarity.

\paragraph{$\alpha$-Precision and $\beta$-Recall.}
These sample-level metrics were introduced by \citet{alaa2022faithful} and we use the implementation provided by Synthcity \citep{qian2023synthcity}. Precision and recall measure fidelity and distributional coverage, respectively. To compute the overall scores, \citet{alaa2022faithful} introduce the notion of $\alpha$-Precision, $P_\alpha$, and $\beta$-Recall, $R_\beta$, defined respectively as
\begin{equation*}
    P_\alpha = P(\vx^s\in \gS_r^\alpha), \text{ for } \alpha \in [0,1], \quad
    R_\beta = P(\vx^r\in\gS_s^\beta), \text{ for } \beta \in [0,1],
\end{equation*}
where $\gS_r^\alpha$ ($\gS_s^\beta$) is a minimum-volume region containing probability mass $\alpha$ ($\beta$) under the real (synthetic) distribution.
Intuitively, $P_\alpha$ corresponds to the fraction of synthetic samples that resemble the most typical fraction $\alpha$ of real samples, while $R_\beta$ is the fraction of real samples covered by the most typical fraction $\beta$ of synthetic samples.

In practice, these quantities are approximated by embedding the real and synthetic data into hyperspheres and computing the fraction of synthetic samples that fall within the $\alpha$-quantile hypersphere ($P_\alpha$) and a similar approach is used to approximate $R_\beta$.
To aggregate the $\alpha$-Precision and $\beta$-Recall values for different values of $\alpha$ and $\beta$ into a single metric, we compute the mean absolute deviation from their optimal value, which represents a straight line with unity slope.
\begin{equation*}
    \Delta P_\alpha = \int_0^1|P_\alpha - \alpha| d\alpha, \qquad \Delta R_\beta = \int_0^1 |R_\beta - \beta| d\beta,
\end{equation*}
which are in practice computed by evaluating the integrands at 30 equally-spaced grid points \citep{qian2023synthcity}. The final scores normalize $\Delta P_\alpha$ and $\Delta R_\beta$ into the range $[0,100]$ with higher values indicating better distributional agreement
\begin{equation*}
    \alpha\text{-Precision} = (1-2\Delta P_\alpha) \times 100 \qquad \beta\text{-Recall} = (1-2\Delta R_\beta) \times 100.
\end{equation*}

\paragraph{Authenticity.}
Authenticity estimates potential memorization using the generator's training rows as the real reference by approximating the probability that the generator  innovates a new sample rather than copies it from the training data (plus some noise).
\citet{alaa2022faithful} propose to compute the Authenticity score as follows. For any synthetic sample $\vx^s$, let $\vx^{r*}(\vx^s) = \argmin_{\vx^r\in\mX^r} d(\vx^s,\vx^r)$ be its closest real sample according to the distance measure $d$. Further, let $D(\vx) = \min_{\vx^r\in\mX^r\setminus\{ \vx \}} d(\vx,\vx^r)$ be the smallest distance between a real sample $\vx$ and any other real sample. Then, the Authenticity score is estimated as
\begin{equation*}
    \mathrm{Authenticity} = 100\times \frac{1}{|\mX^s|} \sum_{\vx^s\in\mX^s} \1[d(\vx^s,\vx^{r*}(\vx^s)) > D(\vx^{r*}(\vx^s))],
\end{equation*}
which effectively computes the fraction of synthetic rows whose distance to the nearest training record exceeds that record's distance to its nearest other training record.

\paragraph{Machine Learning Efficacy (MLE).}
We evaluate the predictive utility of synthetic data by training an XGBoost classifier or regressor on the synthetic dataset and evaluating it on the held-out test dataset. Following TabSyn \citep{zhang2024mixed}, we divide the synthetic dataset into training and validation subsets with an 8:1 ratio and select hyperparameters using validation performance, and the best model is tested on the test dataset mimicking real-world development lifecycles.
For classification datasets, we compute the Area Under the Receiver Operating Characteristic Curve (AUC). For regression datasets, we compute the Normalized Root Mean Squared Error $\mathrm{NRMSE} = \frac{\mathrm{RMSE}}{y_\mathrm{max}-y_\mathrm{min}}$, where $\mathrm{RMSE}$ is computed between predicted and ground-truth targets in the test set and $y_\mathrm{max}$ and $y_\mathrm{min}$ are the largest and smallest target values, respectively.
For all datasets, to obtain a score in $[0,100]$, with higher values indicating better utility, we report
\begin{equation*}
    \mathrm{MLE} = 
    \begin{cases}
        100\times \mathrm{AUC}, & \text{for classification datasets}, \\
        100\times \max(0,1-\mathrm{NRMSE}), & \text{for regression datasets},
    \end{cases}
\end{equation*}
where the $\max$ operator avoids getting a negative score when the $\mathrm{RMSE}$ is higher than the target range.

\paragraph{Distance to closest record (DCR).}
DCR score evaluates whether synthetic records are systematically closer to the generator's training data than to holdout data. It is based on Gower's distance, which is a distance measure between mixed-type objects.
For two mixed-type rows $\vx$ and $\vx'$, this distance is defined as 
\begin{equation*}
    D(\vx,\vx') = \sum_{j\in\gJ_\mathrm{num}} \frac{|x_j-x_j'|}{R_j} + \sum_{j\in\gJ_\mathrm{cat}} \1[x_j\neq x_j'],
\end{equation*}
where $R_j=\max_{\vx\in\mX} x_j - \min_{\vx\in\mX} x_j$ is the range of numerical feature $j$.
For each synthetic record $\vx^s \in\mX_\mathrm{syn}$, we compute its distance to the closest neighbor from the equally-sized train set and from the holdout test set:
\begin{equation*}
    d_T(\vx^s) = \min_{\vx^t\in\mX_\mathrm{train}} D(\vx^s,\vx^t), \qquad d_H(\vx^s) = \min_{\vx^h\in\mX_\mathrm{holdout}} D(\vx^s,\vx^h).
\end{equation*}
The raw score is the fraction of synthetic rows that are closer to the training set than to the holdout set
\begin{equation*}
    q = \frac{1}{N} \sum_{i=1}^N \1[d_T(\vx^s_i)< d_H(\vx^s_i)]
\end{equation*}
Values of $q\leq 0.5$ indicate the absence of excessive proximity to the train set and are preferred. This is reflected in the DCR score normalization
\begin{equation*}
    \mathrm{DCR} = 100\times \min(1,2(1-q)).
\end{equation*}

\subsection{Evaluation datasets}
\label{app:evaluation_datasets}

\begin{table}[t]
\caption{Evaluation dataset statistics. Numerical and categorical column counts include the target. Max Cat denotes the cardinality of the largest categorical vocabulary; Binary denotes binary classification.}
\label{tab:dataset-details}
\begin{center}
\setlength{\tabcolsep}{4pt}
\begin{tabular}{@{}lrrrrrrl@{}}
\toprule
Dataset & \# Train Rows & \# Test Rows & \# Cols & \# Num & \# Cat & \# Max Cat & Task \\
\midrule
Adult & 24,421 & 24,421 & 15 & 6 & 9 & 42 & Binary \\
Default & 15,000 & 15,000 & 24 & 14 & 10 & 11 & Binary \\
Magic & 9,510 & 9,510 & 11 & 10 & 1 & 2 & Binary \\
Shoppers & 6,165 & 6,165 & 18 & 10 & 8 & 20 & Binary \\
Diabetes & 50,883 & 50,883 & 37 & 8 & 29 & 66 & Multiclass \\
Beijing & 20,878 & 20,879 & 8 & 7 & 1 & 4 & Regression \\
News & 19,822 & 19,822 & 47 & 45 & 2 & 7 & Regression \\
\bottomrule
\end{tabular}
\end{center}
\end{table}

We evaluate on the six datasets used by \textsc{TabSyn} \citep{zhang2024mixed}, together with Diabetes, as in \textsc{TabDiff} \citep{shi2025tabdiff}. These datasets cover diverse domains and prediction tasks:
\begin{itemize}
    \item Adult\footnote{\href{https://archive.ics.uci.edu/dataset/2/adult}{https://archive.ics.uci.edu/dataset/2/adult}} (demographics): Census records containing demographic and employment attributes, with a binary classification task predicting whether annual income exceeds USD 50,000.
    \item Default\footnote{\href{https://archive.ics.uci.edu/dataset/350/default+of+credit+card+clients}{https://archive.ics.uci.edu/dataset/350/default+of+credit+card+clients}} (finance): Credit-card client records containing demographic, credit, and repayment information, with a binary classification task predicting payment default.
    \item Magic\footnote{\href{https://archive.ics.uci.edu/dataset/159/magic+gamma+telescope}{https://archive.ics.uci.edu/dataset/159/magic+gamma+telescope}} (astroparticle physics): Simulated telescope observations, with a binary classification task distinguishing gamma-ray events from hadronic background.
    \item Shoppers\footnote{\href{https://archive.ics.uci.edu/dataset/468/online+shoppers+purchasing+intention+dataset}{https://archive.ics.uci.edu/dataset/468/online+shoppers+purchasing+intention+dataset}} (e-commerce): Online browsing-session records, with a binary classification task predicting whether a session results in a purchase.
    \item Diabetes\footnote{\href{https://archive.ics.uci.edu/dataset/296/diabetes+130-us+hospitals+for+years+1999-2008}{https://archive.ics.uci.edu/dataset/296/diabetes+130-us+hospitals+for+years+1999-2008}} (healthcare): Demographic and clinical records of hospital encounters involving patients with diabetes. We retain the original three readmission classes: less than 30 days, more than 30 days, and no readmission, yielding a multiclass classification task.
    \item Beijing\footnote{\href{https://archive.ics.uci.edu/dataset/381/beijing+pm2+5+data}{https://archive.ics.uci.edu/dataset/381/beijing+pm2+5+data}} (environmental monitoring): Air-quality and meteorological measurements, with a regression task predicting PM 2.5 concentrations.
    \item News\footnote{\href{https://archive.ics.uci.edu/dataset/332/online+news+popularity}{https://archive.ics.uci.edu/dataset/332/online+news+popularity}} (online media): Characteristics of online news articles, with a regression task predicting the number of social-media shares.
\end{itemize}

Table~\ref{tab:dataset-details} summarizes the dataset sizes, feature composition, categorical cardinalities, and prediction tasks.

\subsection{Pre-training datasets}
\label{app:pre_training_datasets}

\begin{table}[h]
\caption{Pretraining dataset summary across the 337 datasets.}
\label{tab:pretraining-dataset-summary}
\begin{center}
\setlength{\tabcolsep}{4pt}
\begin{tabular}{@{}lrrrrr@{}}
\toprule
Quantity & Mean & Median & Min & Max & Sum \\
\midrule
\# Rows & 28,585 & 10,999 & 10 & 100,000 & 9,633,139 \\
\# Cols & 11.1 & 10 & 2 & 33 & 3,734 \\
\# Num & 7.3 & 6 & 0 & 28 & 2,458 \\
\# Cat & 3.8 & 2 & 0 & 23 & 1,276 \\
\midrule
\multicolumn{6}{@{}l@{}}{\textbf{Total datasets:} 337} \\
\bottomrule
\end{tabular}
\end{center}
\end{table}

\begin{figure}[t]
    \begin{center}
        \includegraphics[width=0.6\linewidth]{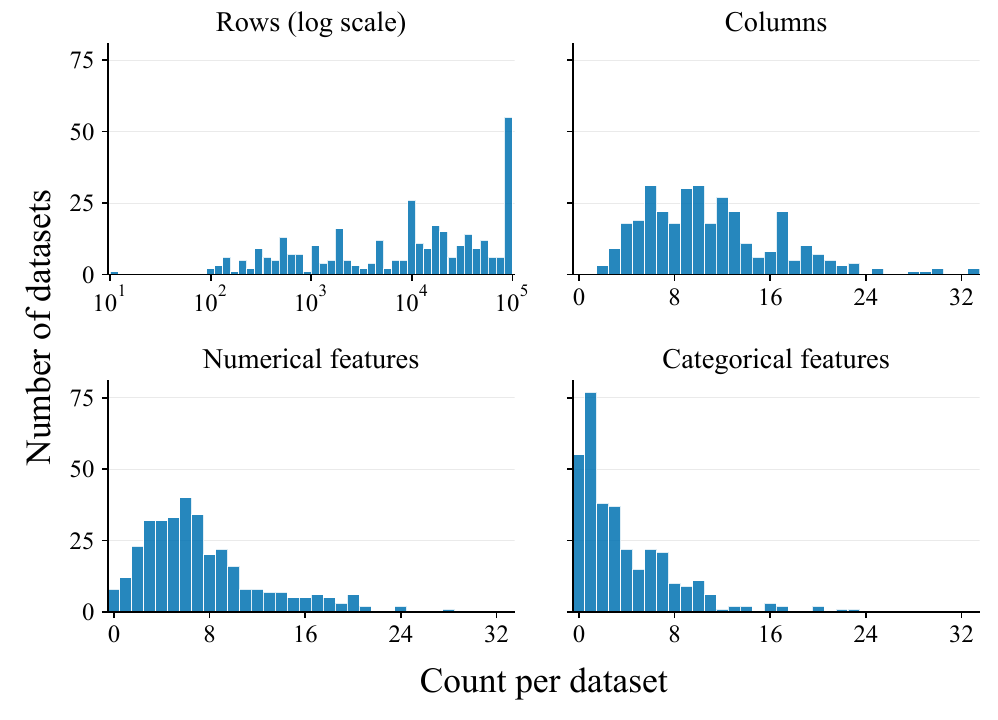}
    \end{center}
    \caption{
        Distribution of row counts and total, numerical, and categorical
        column counts across the 337 pre-training datasets.
        Each dataset contributes one observation, and column counts reflect
        retained features after preprocessing.
        Row counts use 50 logarithmically spaced bins; column counts use
        one bin per integer.
    }
    \label{fig:pretraining-dataset-distributions}
\end{figure}

We construct our pre-training corpus from the datasets curated by \citet{yan2024making}, drawn from TabPertNet, now known as OpenTabs \citep{ye2024towards}. This source collection contains heterogeneous tables from OpenML, UCI, Kaggle, and Data.gov, with numerical and categorical features. The curated subset covers binary classification and regression tasks and excludes datasets with uninformative column names or categorical codes that cannot be mapped to meaningful values. The same source collection also underlies the pre-training data used by \textsc{CTSyn} \citep{lin2025ctsyn}.

To identify potential overlap with our evaluation datasets, we automatically compare dataset names, column names, marginal distributions, and categorical values between each evaluation dataset and every dataset in the pre-training pool. We then manually inspect the flagged datasets and remove those identified as copies or subsets of the evaluation datasets. This procedure removes 10 of the initial 347 datasets, leaving 337 datasets for pre-training. Table~\ref{tab:pretraining-dataset-summary} and Figure~\ref{fig:pretraining-dataset-distributions} summarize the resulting corpus.

\subsection{Implementation Details}
\label{app:implementation_details}

\paragraph{Data pre-processing.}
We impute missing numerical values with the corresponding training column mean and represent missing categorical values with a dedicated category. To reduce differences in scale and marginal distributions across numerical features, we apply a quantile transformation. Quantile transformations are fitted exclusively on the training data; in few-shot settings, only the $N$ available training samples are used for fitting. After sampling, we apply the corresponding inverse transformations to map generated numerical values back to their original scale.

\paragraph{Baselines.}
We train and evaluate all baselines under the same experimental protocol as our model, using identical data splits, training subsets, and evaluation metrics. For each baseline, we use its official implementation with the default hyperparameters, without additional tuning. Since our evaluation datasets follow established benchmarks used in prior work, we retain the configurations provided by the original authors.
For two-stage models, we count one epoch as a complete pass over the data by the autoencoder and the latent diffusion model.

\paragraph{\ours{} hyperparameters.}
We optimize our model using \textsc{AdamW} with a learning rate of $10^{-4}$ for 2,000 epochs, using a batch size of 4096. We apply a linear learning-rate warmup over the first 5\% of training steps, and subsequently multiply the learning rate by 0.9 whenever the training loss does not improve for 50 consecutive epochs. During few-shot fine-tuning, we freeze the transformer encoder. We generate synthetic samples using 50 sampling steps on a linear time grid.
In the few-epoch adaptation experiment, we do not apply the learning-rate warmup stage and train for the specified number of epochs.

\section{Detailed Results}
\label{app:detailed_results}

In this section, we provide detailed evaluation results for our experiments presented in Section~\ref{sec:experiments}.

\subsection{Cross-Dataset Generation}
\label{app:detailed_cross_dataset}

For the in-domain cross-dataset generation experiments (Section~\ref{sec:exp:cross_dataset_generation}), we present per-axis evaluation results in Table~\ref{tab:fidelity__utility__privacy}. All four fidelity metrics are in Table~\ref{tab:shape__trend__precision__recall} and the two privacy metrics are in Table~\ref{tab:dcr_score__authenticity}. Since Utility has only one metric, the metric score is the same as the aggregate score presented in Table~\ref{tab:fidelity__utility__privacy}.

\begin{table}[t]
\caption{In-domain cross-dataset generation results for the three evaluation axes}
\label{tab:fidelity__utility__privacy}
\centering
\small
\resizebox{\linewidth}{!}{%
\begin{tabular}{@{}c@{\hspace{0.4em}}l|c|c|c|c|c|c|c|c}
\toprule
\multicolumn{10}{c}{Fidelity} \\
\midrule
 & {\renewcommand{\arraystretch}{0.6}\diagbox[font={\fontsize{5.5}{6.6}\selectfont},width=8.5em,height=1.1em]{Method}{Dataset}} & Adult & Default & Magic & Shoppers & Diabetes & Beijing & News & Average \\
\midrule
 & \textsc{Train Set} & \(86.52\,{\scriptstyle \pm 0.00}\) & \(87.06\,{\scriptstyle \pm 0.00}\) & \(86.61\,{\scriptstyle \pm 0.00}\) & \(86.37\,{\scriptstyle \pm 0.00}\) & \(86.84\,{\scriptstyle \pm 0.00}\) & \(87.21\,{\scriptstyle \pm 0.00}\) & \(85.07\,{\scriptstyle \pm 0.00}\) & \(86.53\) \\
\midrule[0.2pt]
\multirow{7}{*}[0pt]{\rotatebox[origin=c]{90}{\fontsize{5.5}{6.6}\selectfont Single-dataset}} & \textsc{SMOTE} & \(82.35\,{\scriptstyle \pm 0.10}\) & \(84.31\,{\scriptstyle \pm 0.16}\) & \(84.49\,{\scriptstyle \pm 0.26}\) & \(82.26\,{\scriptstyle \pm 0.19}\) & \(81.05\,{\scriptstyle \pm 0.08}\) & \(\mathbf{84.75}\,{\scriptstyle \pm 0.04}\) & \(\underline{81.39}\,{\scriptstyle \pm 0.80}\) & \(82.94\) \\
 & \textsc{CTGAN} & \(67.62\,{\scriptstyle \pm 2.70}\) & \(64.91\,{\scriptstyle \pm 1.55}\) & \(66.43\,{\scriptstyle \pm 1.12}\) & \(67.75\,{\scriptstyle \pm 1.22}\) & \(64.35\,{\scriptstyle \pm 4.79}\) & \(75.59\,{\scriptstyle \pm 1.02}\) & \(66.04\,{\scriptstyle \pm 0.66}\) & \(67.53\) \\
 & \textsc{TVAE} & \(76.38\,{\scriptstyle \pm 1.56}\) & \(73.32\,{\scriptstyle \pm 1.12}\) & \(76.98\,{\scriptstyle \pm 0.39}\) & \(57.64\,{\scriptstyle \pm 0.73}\) & \(60.59\,{\scriptstyle \pm 1.65}\) & \(76.11\,{\scriptstyle \pm 1.65}\) & \(73.12\,{\scriptstyle \pm 0.51}\) & \(70.59\) \\
 & \textsc{GReaT} & \(74.05\,{\scriptstyle \pm 0.14}\) & \(77.30\,{\scriptstyle \pm 0.32}\) & \(74.07\,{\scriptstyle \pm 0.40}\) & \(75.88\,{\scriptstyle \pm 0.29}\) & \(73.92\,{\scriptstyle \pm 0.11}\) & \(80.01\,{\scriptstyle \pm 0.08}\) & -- & \(75.87\) \\
 & \textsc{TabDDPM} & \(84.71\,{\scriptstyle \pm 0.10}\) & \(83.12\,{\scriptstyle \pm 0.34}\) & \(84.69\,{\scriptstyle \pm 0.17}\) & \(80.42\,{\scriptstyle \pm 0.14}\) & \(32.44\,{\scriptstyle \pm 0.00}\) & \(83.09\,{\scriptstyle \pm 0.07}\) & \(18.83\,{\scriptstyle \pm 0.02}\) & \(66.76\) \\
 & \textsc{TabSyn} & \(\underline{85.81}\,{\scriptstyle \pm 0.02}\) & \(85.38\,{\scriptstyle \pm 0.07}\) & \(\underline{85.58}\,{\scriptstyle \pm 0.10}\) & \(\underline{84.54}\,{\scriptstyle \pm 0.39}\) & \(80.28\,{\scriptstyle \pm 0.00}\) & \(83.23\,{\scriptstyle \pm 0.10}\) & \(80.14\,{\scriptstyle \pm 0.13}\) & \(\underline{83.57}\) \\
 & \textsc{TabDiff} & \(85.53\,{\scriptstyle \pm 0.02}\) & \(\underline{85.68}\,{\scriptstyle \pm 0.03}\) & \(85.53\,{\scriptstyle \pm 0.09}\) & \(84.18\,{\scriptstyle \pm 0.37}\) & \(\underline{84.25}\,{\scriptstyle \pm 0.04}\) & \(\underline{83.44}\,{\scriptstyle \pm 0.09}\) & \(73.66\,{\scriptstyle \pm 0.08}\) & \(83.18\) \\
\midrule[0.2pt]
\multirow{3}{*}[0pt]{\rotatebox[origin=c]{90}{\fontsize{5.5}{6.6}\selectfont Cross-dataset}} & \textsc{TabEBM} & \(38.07\,{\scriptstyle \pm 0.08}\) & \(50.41\,{\scriptstyle \pm 0.20}\) & \(78.75\,{\scriptstyle \pm 0.04}\) & \(51.66\,{\scriptstyle \pm 0.19}\) & \(29.86\,{\scriptstyle \pm 0.28}\) & \(68.01\,{\scriptstyle \pm 0.15}\) & \(53.77\,{\scriptstyle \pm 11.75}\) & \(52.93\) \\
 & \textsc{CTSyn} & \(58.68\,{\scriptstyle \pm 0.10}\) & \(72.77\,{\scriptstyle \pm 0.11}\) & \(66.30\,{\scriptstyle \pm 0.02}\) & \(65.52\,{\scriptstyle \pm 0.14}\) & \(43.55\,{\scriptstyle \pm 0.02}\) & \(66.93\,{\scriptstyle \pm 0.10}\) & \(70.61\,{\scriptstyle \pm 0.07}\) & \(63.48\) \\
 & \textsc{CDMD (Ours)} & \(\mathbf{85.97}\,{\scriptstyle \pm 0.02}\) & \(\mathbf{86.08}\,{\scriptstyle \pm 0.07}\) & \(\mathbf{85.66}\,{\scriptstyle \pm 0.16}\) & \(\mathbf{85.39}\,{\scriptstyle \pm 0.13}\) & \(\mathbf{85.60}\,{\scriptstyle \pm 0.07}\) & \(83.36\,{\scriptstyle \pm 0.02}\) & \(\mathbf{82.47}\,{\scriptstyle \pm 0.06}\) & \(\mathbf{84.93}\) \\
\midrule[\heavyrulewidth]
\multicolumn{10}{c}{Utility} \\
\midrule
 & {\renewcommand{\arraystretch}{0.6}\diagbox[font={\fontsize{5.5}{6.6}\selectfont},width=8.5em,height=1.1em]{Method}{Dataset}} & Adult & Default & Magic & Shoppers & Diabetes & Beijing & News & Average \\
\midrule
 & \textsc{Train Set} & \(92.75\,{\scriptstyle \pm 0.00}\) & \(77.98\,{\scriptstyle \pm 0.00}\) & \(92.78\,{\scriptstyle \pm 0.00}\) & \(92.76\,{\scriptstyle \pm 0.00}\) & \(67.97\,{\scriptstyle \pm 0.00}\) & \(89.09\,{\scriptstyle \pm 0.00}\) & \(89.86\,{\scriptstyle \pm 0.00}\) & \(86.17\) \\
\midrule[0.2pt]
\multirow{7}{*}[0pt]{\rotatebox[origin=c]{90}{\fontsize{5.5}{6.6}\selectfont Single-dataset}} & \textsc{SMOTE} & \(90.00\,{\scriptstyle \pm 0.11}\) & \(74.35\,{\scriptstyle \pm 0.25}\) & \(\mathbf{92.32}\,{\scriptstyle \pm 0.04}\) & \(91.20\,{\scriptstyle \pm 0.24}\) & \(63.78\,{\scriptstyle \pm 0.05}\) & \(88.70\,{\scriptstyle \pm 0.06}\) & \(\mathbf{89.92}\,{\scriptstyle \pm 0.04}\) & \(84.32\) \\
 & \textsc{CTGAN} & \(88.49\,{\scriptstyle \pm 0.20}\) & \(71.96\,{\scriptstyle \pm 0.73}\) & \(82.68\,{\scriptstyle \pm 2.44}\) & \(81.64\,{\scriptstyle \pm 3.94}\) & \(55.57\,{\scriptstyle \pm 0.89}\) & \(86.97\,{\scriptstyle \pm 0.36}\) & \(89.24\,{\scriptstyle \pm 0.13}\) & \(79.51\) \\
 & \textsc{TVAE} & \(88.57\,{\scriptstyle \pm 0.22}\) & \(74.99\,{\scriptstyle \pm 0.33}\) & \(83.53\,{\scriptstyle \pm 2.39}\) & \(88.20\,{\scriptstyle \pm 0.38}\) & \(60.46\,{\scriptstyle \pm 1.16}\) & \(87.11\,{\scriptstyle \pm 0.27}\) & \(87.78\,{\scriptstyle \pm 0.17}\) & \(81.52\) \\
 & \textsc{GReaT} & \(\mathbf{91.22}\,{\scriptstyle \pm 0.29}\) & \(75.79\,{\scriptstyle \pm 0.79}\) & \(90.01\,{\scriptstyle \pm 0.14}\) & \(91.03\,{\scriptstyle \pm 0.43}\) & \(\mathbf{66.73}\,{\scriptstyle \pm 0.34}\) & \(88.28\,{\scriptstyle \pm 0.20}\) & -- & \(83.84\) \\
 & \textsc{TabDDPM} & \(90.81\,{\scriptstyle \pm 0.11}\) & \(76.58\,{\scriptstyle \pm 0.46}\) & \(91.97\,{\scriptstyle \pm 0.06}\) & \(90.37\,{\scriptstyle \pm 0.17}\) & \(50.51\,{\scriptstyle \pm 0.36}\) & \(\underline{88.92}\,{\scriptstyle \pm 0.56}\) & \(58.35\,{\scriptstyle \pm 11.80}\) & \(78.21\) \\
 & \textsc{TabSyn} & \(91.05\,{\scriptstyle \pm 0.14}\) & \(\mathbf{77.34}\,{\scriptstyle \pm 0.23}\) & \(92.11\,{\scriptstyle \pm 0.10}\) & \(91.15\,{\scriptstyle \pm 0.53}\) & \(64.38\,{\scriptstyle \pm 0.17}\) & \(\mathbf{89.13}\,{\scriptstyle \pm 0.24}\) & \(89.60\,{\scriptstyle \pm 0.30}\) & \(84.97\) \\
 & \textsc{TabDiff} & \(91.04\,{\scriptstyle \pm 0.18}\) & \(77.09\,{\scriptstyle \pm 0.44}\) & \(91.82\,{\scriptstyle \pm 0.08}\) & \(\mathbf{91.68}\,{\scriptstyle \pm 0.22}\) & \(66.50\,{\scriptstyle \pm 0.31}\) & \(88.86\,{\scriptstyle \pm 0.53}\) & \(\underline{89.85}\,{\scriptstyle \pm 0.27}\) & \(\mathbf{85.26}\) \\
\midrule[0.2pt]
\multirow{3}{*}[0pt]{\rotatebox[origin=c]{90}{\fontsize{5.5}{6.6}\selectfont Cross-dataset}} & \textsc{TabEBM} & \(86.73\,{\scriptstyle \pm 1.24}\) & \(64.40\,{\scriptstyle \pm 1.80}\) & \(89.75\,{\scriptstyle \pm 0.37}\) & \(88.33\,{\scriptstyle \pm 1.22}\) & \(57.07\,{\scriptstyle \pm 0.55}\) & \(86.61\,{\scriptstyle \pm 0.36}\) & \(81.56\,{\scriptstyle \pm 0.58}\) & \(79.21\) \\
 & \textsc{CTSyn} & \(76.63\,{\scriptstyle \pm 1.14}\) & \(60.82\,{\scriptstyle \pm 1.48}\) & \(78.63\,{\scriptstyle \pm 0.71}\) & \(61.82\,{\scriptstyle \pm 3.46}\) & \(52.52\,{\scriptstyle \pm 0.06}\) & \(85.20\,{\scriptstyle \pm 0.18}\) & \(88.70\,{\scriptstyle \pm 0.19}\) & \(72.05\) \\
 & \textsc{CDMD (Ours)} & \(\underline{91.21}\,{\scriptstyle \pm 0.07}\) & \(\underline{77.27}\,{\scriptstyle \pm 0.21}\) & \(\underline{92.18}\,{\scriptstyle \pm 0.09}\) & \(\underline{91.58}\,{\scriptstyle \pm 0.15}\) & \(\underline{66.58}\,{\scriptstyle \pm 0.04}\) & \(88.43\,{\scriptstyle \pm 0.52}\) & \(89.54\,{\scriptstyle \pm 0.02}\) & \(\underline{85.26}\) \\
\midrule[\heavyrulewidth]
\multicolumn{10}{c}{Privacy} \\
\midrule
 & {\renewcommand{\arraystretch}{0.6}\diagbox[font={\fontsize{5.5}{6.6}\selectfont},width=8.5em,height=1.1em]{Method}{Dataset}} & Adult & Default & Magic & Shoppers & Diabetes & Beijing & News & Average \\
\midrule
 & \textsc{Train Set} & \(0.09\,{\scriptstyle \pm 0.00}\) & \(0.11\,{\scriptstyle \pm 0.00}\) & \(0.60\,{\scriptstyle \pm 0.00}\) & \(1.14\,{\scriptstyle \pm 0.00}\) & \(0.00\,{\scriptstyle \pm 0.00}\) & \(0.01\,{\scriptstyle \pm 0.00}\) & \(0.00\,{\scriptstyle \pm 0.00}\) & \(0.28\) \\
\midrule[0.2pt]
\multirow{7}{*}[0pt]{\rotatebox[origin=c]{90}{\fontsize{5.5}{6.6}\selectfont Single-dataset}} & \textsc{SMOTE} & \(23.55\,{\scriptstyle \pm 0.18}\) & \(23.71\,{\scriptstyle \pm 0.17}\) & \(16.99\,{\scriptstyle \pm 0.27}\) & \(16.93\,{\scriptstyle \pm 0.65}\) & \(12.58\,{\scriptstyle \pm 0.07}\) & \(26.42\,{\scriptstyle \pm 0.15}\) & \(14.35\,{\scriptstyle \pm 0.16}\) & \(19.22\) \\
 & \textsc{CTGAN} & \(78.91\,{\scriptstyle \pm 0.28}\) & \(79.75\,{\scriptstyle \pm 0.63}\) & \(\underline{87.67}\,{\scriptstyle \pm 0.14}\) & \(80.31\,{\scriptstyle \pm 0.26}\) & \(86.71\,{\scriptstyle \pm 1.01}\) & \(\underline{85.97}\,{\scriptstyle \pm 0.33}\) & \(\underline{85.51}\,{\scriptstyle \pm 0.45}\) & \(\underline{83.55}\) \\
 & \textsc{TVAE} & \(77.46\,{\scriptstyle \pm 0.51}\) & \(\underline{81.76}\,{\scriptstyle \pm 1.16}\) & \(81.41\,{\scriptstyle \pm 1.28}\) & \(\mathbf{87.05}\,{\scriptstyle \pm 2.03}\) & \(82.35\,{\scriptstyle \pm 0.51}\) & \(84.58\,{\scriptstyle \pm 0.87}\) & \(81.29\,{\scriptstyle \pm 0.28}\) & \(82.27\) \\
 & \textsc{GReaT} & \(73.20\,{\scriptstyle \pm 0.16}\) & \(75.13\,{\scriptstyle \pm 0.45}\) & \(76.84\,{\scriptstyle \pm 0.51}\) & \(76.04\,{\scriptstyle \pm 0.48}\) & \(74.21\,{\scriptstyle \pm 0.22}\) & \(73.25\,{\scriptstyle \pm 0.15}\) & -- & \(74.78\) \\
 & \textsc{TabDDPM} & \(74.51\,{\scriptstyle \pm 0.33}\) & \(74.68\,{\scriptstyle \pm 0.07}\) & \(72.88\,{\scriptstyle \pm 0.35}\) & \(65.11\,{\scriptstyle \pm 0.32}\) & \(\mathbf{98.65}\,{\scriptstyle \pm 0.17}\) & \(81.28\,{\scriptstyle \pm 0.26}\) & \(\mathbf{100.00}\,{\scriptstyle \pm 0.00}\) & \(81.02\) \\
 & \textsc{TabSyn} & \(73.97\,{\scriptstyle \pm 0.10}\) & \(74.28\,{\scriptstyle \pm 0.28}\) & \(74.27\,{\scriptstyle \pm 0.30}\) & \(74.21\,{\scriptstyle \pm 0.88}\) & \(78.47\,{\scriptstyle \pm 0.19}\) & \(80.61\,{\scriptstyle \pm 0.13}\) & \(77.41\,{\scriptstyle \pm 0.25}\) & \(76.18\) \\
 & \textsc{TabDiff} & \(66.89\,{\scriptstyle \pm 0.19}\) & \(66.91\,{\scriptstyle \pm 0.32}\) & \(73.61\,{\scriptstyle \pm 0.31}\) & \(74.43\,{\scriptstyle \pm 0.21}\) & \(67.64\,{\scriptstyle \pm 0.35}\) & \(80.58\,{\scriptstyle \pm 0.31}\) & \(74.78\,{\scriptstyle \pm 0.13}\) & \(72.12\) \\
\midrule[0.2pt]
\multirow{3}{*}[0pt]{\rotatebox[origin=c]{90}{\fontsize{5.5}{6.6}\selectfont Cross-dataset}} & \textsc{TabEBM} & \(\underline{83.85}\,{\scriptstyle \pm 0.37}\) & \(\mathbf{89.15}\,{\scriptstyle \pm 0.66}\) & \(40.59\,{\scriptstyle \pm 0.21}\) & \(75.87\,{\scriptstyle \pm 0.70}\) & \(\underline{93.43}\,{\scriptstyle \pm 0.04}\) & \(82.30\,{\scriptstyle \pm 0.30}\) & \(53.62\,{\scriptstyle \pm 0.31}\) & \(74.12\) \\
 & \textsc{CTSyn} & \(\mathbf{84.48}\,{\scriptstyle \pm 0.07}\) & \(81.56\,{\scriptstyle \pm 0.35}\) & \(\mathbf{89.29}\,{\scriptstyle \pm 0.72}\) & \(\underline{81.27}\,{\scriptstyle \pm 1.08}\) & \(92.35\,{\scriptstyle \pm 0.05}\) & \(\mathbf{96.46}\,{\scriptstyle \pm 0.07}\) & \(79.49\,{\scriptstyle \pm 0.40}\) & \(\mathbf{86.41}\) \\
 & \textsc{CDMD (Ours)} & \(74.12\,{\scriptstyle \pm 0.50}\) & \(74.00\,{\scriptstyle \pm 0.79}\) & \(74.13\,{\scriptstyle \pm 0.70}\) & \(72.40\,{\scriptstyle \pm 0.24}\) & \(75.34\,{\scriptstyle \pm 0.15}\) & \(80.45\,{\scriptstyle \pm 0.19}\) & \(75.35\,{\scriptstyle \pm 0.19}\) & \(75.11\) \\
\bottomrule
\end{tabular}
}%
\end{table}

\begin{table}[t]
\caption{In-domain cross-dataset generation results for the four individual metrics of Fidelity.}
\label{tab:shape__trend__precision__recall}
\centering
\small
\resizebox{\linewidth}{!}{%
\begin{tabular}{@{}c@{\hspace{0.4em}}l|c|c|c|c|c|c|c|c}
\toprule
\multicolumn{10}{c}{\textbf{Shape}} \\
\midrule
 & {\renewcommand{\arraystretch}{0.6}\diagbox[font={\fontsize{5.5}{6.6}\selectfont},width=8.5em,height=1.1em]{Method}{Dataset}} & Adult & Default & Magic & Shoppers & Diabetes & Beijing & News & Average \\
\midrule
 & \textsc{Train Set} & \(99.38\,{\scriptstyle \pm 0.00}\) & \(99.23\,{\scriptstyle \pm 0.00}\) & \(99.08\,{\scriptstyle \pm 0.00}\) & \(98.84\,{\scriptstyle \pm 0.00}\) & \(99.73\,{\scriptstyle \pm 0.00}\) & \(99.32\,{\scriptstyle \pm 0.00}\) & \(99.20\,{\scriptstyle \pm 0.00}\) & \(99.25\) \\
\midrule[0.2pt]
\multirow{7}{*}[0pt]{\rotatebox[origin=c]{90}{\fontsize{5.5}{6.6}\selectfont Single-dataset}} & \textsc{SMOTE} & \(98.22\,{\scriptstyle \pm 0.03}\) & \(97.99\,{\scriptstyle \pm 0.10}\) & \(98.47\,{\scriptstyle \pm 0.08}\) & \(97.48\,{\scriptstyle \pm 0.04}\) & \(98.48\,{\scriptstyle \pm 0.01}\) & \(\mathbf{99.18}\,{\scriptstyle \pm 0.10}\) & \(96.28\,{\scriptstyle \pm 0.04}\) & \(98.02\) \\
 & \textsc{CTGAN} & \(85.73\,{\scriptstyle \pm 2.07}\) & \(85.51\,{\scriptstyle \pm 0.64}\) & \(89.27\,{\scriptstyle \pm 1.10}\) & \(81.14\,{\scriptstyle \pm 1.38}\) & \(89.91\,{\scriptstyle \pm 0.50}\) & \(92.52\,{\scriptstyle \pm 1.50}\) & \(79.45\,{\scriptstyle \pm 0.96}\) & \(86.22\) \\
 & \textsc{TVAE} & \(89.32\,{\scriptstyle \pm 1.38}\) & \(89.85\,{\scriptstyle \pm 0.33}\) & \(92.45\,{\scriptstyle \pm 0.75}\) & \(77.50\,{\scriptstyle \pm 0.36}\) & \(91.44\,{\scriptstyle \pm 0.72}\) & \(93.82\,{\scriptstyle \pm 1.92}\) & \(82.50\,{\scriptstyle \pm 0.29}\) & \(88.13\) \\
 & \textsc{GReaT} & \(91.86\,{\scriptstyle \pm 0.08}\) & \(85.88\,{\scriptstyle \pm 0.27}\) & \(83.92\,{\scriptstyle \pm 0.06}\) & \(85.74\,{\scriptstyle \pm 0.10}\) & \(93.83\,{\scriptstyle \pm 0.02}\) & \(89.57\,{\scriptstyle \pm 0.12}\) & -- & \(88.47\) \\
 & \textsc{TabDDPM} & \(98.71\,{\scriptstyle \pm 0.03}\) & \(95.71\,{\scriptstyle \pm 0.11}\) & \(98.24\,{\scriptstyle \pm 0.16}\) & \(95.16\,{\scriptstyle \pm 0.07}\) & \(70.27\,{\scriptstyle \pm 0.01}\) & \(98.88\,{\scriptstyle \pm 0.12}\) & \(14.05\,{\scriptstyle \pm 0.07}\) & \(81.57\) \\
 & \textsc{TabSyn} & \(\underline{99.09}\,{\scriptstyle \pm 0.03}\) & \(\underline{98.59}\,{\scriptstyle \pm 0.03}\) & \(98.56\,{\scriptstyle \pm 0.08}\) & \(97.62\,{\scriptstyle \pm 0.16}\) & \(97.92\,{\scriptstyle \pm 0.02}\) & \(98.94\,{\scriptstyle \pm 0.04}\) & \(95.91\,{\scriptstyle \pm 0.02}\) & \(98.09\) \\
 & \textsc{TabDiff} & \(98.98\,{\scriptstyle \pm 0.10}\) & \(98.39\,{\scriptstyle \pm 0.10}\) & \(\mathbf{98.76}\,{\scriptstyle \pm 0.05}\) & \(\underline{97.94}\,{\scriptstyle \pm 0.10}\) & \(\underline{99.05}\,{\scriptstyle \pm 0.03}\) & \(99.00\,{\scriptstyle \pm 0.14}\) & \(\underline{96.77}\,{\scriptstyle \pm 0.01}\) & \(\underline{98.41}\) \\
\midrule[0.2pt]
\multirow{3}{*}[0pt]{\rotatebox[origin=c]{90}{\fontsize{5.5}{6.6}\selectfont Cross-dataset}} & \textsc{TabEBM} & \(70.86\,{\scriptstyle \pm 0.03}\) & \(75.57\,{\scriptstyle \pm 0.37}\) & \(96.06\,{\scriptstyle \pm 0.00}\) & \(72.98\,{\scriptstyle \pm 0.04}\) & \(69.21\,{\scriptstyle \pm 0.77}\) & \(79.21\,{\scriptstyle \pm 0.44}\) & \(82.74\,{\scriptstyle \pm 0.79}\) & \(78.09\) \\
 & \textsc{CTSyn} & \(89.77\,{\scriptstyle \pm 0.03}\) & \(91.30\,{\scriptstyle \pm 0.05}\) & \(84.63\,{\scriptstyle \pm 0.07}\) & \(88.16\,{\scriptstyle \pm 0.03}\) & \(88.73\,{\scriptstyle \pm 0.02}\) & \(87.76\,{\scriptstyle \pm 0.02}\) & \(89.27\,{\scriptstyle \pm 0.07}\) & \(88.52\) \\
 & \textsc{CDMD (Ours)} & \(\mathbf{99.18}\,{\scriptstyle \pm 0.10}\) & \(\mathbf{98.83}\,{\scriptstyle \pm 0.07}\) & \(\underline{98.70}\,{\scriptstyle \pm 0.20}\) & \(\mathbf{98.47}\,{\scriptstyle \pm 0.01}\) & \(\mathbf{99.54}\,{\scriptstyle \pm 0.02}\) & \(\underline{99.07}\,{\scriptstyle \pm 0.07}\) & \(\mathbf{98.48}\,{\scriptstyle \pm 0.02}\) & \(\mathbf{98.90}\) \\
\midrule[\heavyrulewidth]
\multicolumn{10}{c}{\textbf{Trend}} \\
\midrule
 & {\renewcommand{\arraystretch}{0.6}\diagbox[font={\fontsize{5.5}{6.6}\selectfont},width=8.5em,height=1.1em]{Method}{Dataset}} & Adult & Default & Magic & Shoppers & Diabetes & Beijing & News & Average \\
\midrule
 & \textsc{Train Set} & \(98.65\,{\scriptstyle \pm 0.00}\) & \(98.95\,{\scriptstyle \pm 0.00}\) & \(97.60\,{\scriptstyle \pm 0.00}\) & \(98.47\,{\scriptstyle \pm 0.00}\) & \(99.30\,{\scriptstyle \pm 0.00}\) & \(99.96\,{\scriptstyle \pm 0.00}\) & \(91.90\,{\scriptstyle \pm 0.00}\) & \(97.83\) \\
\midrule[0.2pt]
\multirow{7}{*}[0pt]{\rotatebox[origin=c]{90}{\fontsize{5.5}{6.6}\selectfont Single-dataset}} & \textsc{SMOTE} & \(97.19\,{\scriptstyle \pm 0.01}\) & \(97.19\,{\scriptstyle \pm 0.08}\) & \(95.97\,{\scriptstyle \pm 0.84}\) & \(\underline{97.17}\,{\scriptstyle \pm 0.11}\) & \(97.19\,{\scriptstyle \pm 0.10}\) & \(99.37\,{\scriptstyle \pm 0.05}\) & \(\mathbf{92.29}\,{\scriptstyle \pm 3.38}\) & \(96.63\) \\
 & \textsc{CTGAN} & \(80.80\,{\scriptstyle \pm 2.16}\) & \(82.67\,{\scriptstyle \pm 1.23}\) & \(86.10\,{\scriptstyle \pm 0.55}\) & \(80.81\,{\scriptstyle \pm 0.10}\) & \(80.95\,{\scriptstyle \pm 1.91}\) & \(98.03\,{\scriptstyle \pm 0.64}\) & \(83.31\,{\scriptstyle \pm 0.82}\) & \(84.67\) \\
 & \textsc{TVAE} & \(89.86\,{\scriptstyle \pm 0.56}\) & \(93.86\,{\scriptstyle \pm 0.17}\) & \(91.26\,{\scriptstyle \pm 1.65}\) & \(89.97\,{\scriptstyle \pm 0.89}\) & \(93.52\,{\scriptstyle \pm 1.77}\) & \(95.56\,{\scriptstyle \pm 0.54}\) & \(90.13\,{\scriptstyle \pm 0.49}\) & \(92.02\) \\
 & \textsc{GReaT} & \(88.34\,{\scriptstyle \pm 0.20}\) & \(90.54\,{\scriptstyle \pm 0.30}\) & \(88.59\,{\scriptstyle \pm 1.46}\) & \(91.52\,{\scriptstyle \pm 1.06}\) & \(87.88\,{\scriptstyle \pm 0.19}\) & \(98.13\,{\scriptstyle \pm 0.01}\) & -- & \(90.83\) \\
 & \textsc{TabDDPM} & \(97.25\,{\scriptstyle \pm 0.08}\) & \(95.13\,{\scriptstyle \pm 0.13}\) & \(96.41\,{\scriptstyle \pm 0.30}\) & \(87.10\,{\scriptstyle \pm 0.07}\) & \(48.35\,{\scriptstyle \pm 0.01}\) & \(99.36\,{\scriptstyle \pm 0.05}\) & \(61.28\,{\scriptstyle \pm 0.09}\) & \(83.55\) \\
 & \textsc{TabSyn} & \(\underline{97.83}\,{\scriptstyle \pm 0.03}\) & \(98.19\,{\scriptstyle \pm 0.08}\) & \(\underline{97.10}\,{\scriptstyle \pm 0.16}\) & \(96.86\,{\scriptstyle \pm 0.24}\) & \(96.01\,{\scriptstyle \pm 0.12}\) & \(\underline{99.92}\,{\scriptstyle \pm 0.02}\) & \(91.48\,{\scriptstyle \pm 0.32}\) & \(\underline{96.77}\) \\
 & \textsc{TabDiff} & \(97.80\,{\scriptstyle \pm 0.12}\) & \(\underline{98.29}\,{\scriptstyle \pm 0.06}\) & \(97.04\,{\scriptstyle \pm 0.36}\) & \(96.48\,{\scriptstyle \pm 0.27}\) & \(\underline{98.49}\,{\scriptstyle \pm 0.08}\) & \(\mathbf{99.94}\,{\scriptstyle \pm 0.00}\) & \(84.19\,{\scriptstyle \pm 0.14}\) & \(96.03\) \\
\midrule[0.2pt]
\multirow{3}{*}[0pt]{\rotatebox[origin=c]{90}{\fontsize{5.5}{6.6}\selectfont Cross-dataset}} & \textsc{TabEBM} & \(55.93\,{\scriptstyle \pm 0.10}\) & \(73.00\,{\scriptstyle \pm 0.03}\) & \(91.71\,{\scriptstyle \pm 0.13}\) & \(73.21\,{\scriptstyle \pm 0.19}\) & \(48.71\,{\scriptstyle \pm 0.02}\) & \(99.22\,{\scriptstyle \pm 0.03}\) & \(88.02\,{\scriptstyle \pm 3.98}\) & \(75.68\) \\
 & \textsc{CTSyn} & \(73.59\,{\scriptstyle \pm 0.12}\) & \(91.02\,{\scriptstyle \pm 0.06}\) & \(82.08\,{\scriptstyle \pm 0.01}\) & \(82.05\,{\scriptstyle \pm 0.19}\) & \(73.98\,{\scriptstyle \pm 0.06}\) & \(91.21\,{\scriptstyle \pm 0.01}\) & \(87.58\,{\scriptstyle \pm 0.22}\) & \(83.07\) \\
 & \textsc{CDMD (Ours)} & \(\mathbf{98.18}\,{\scriptstyle \pm 0.06}\) & \(\mathbf{98.58}\,{\scriptstyle \pm 0.05}\) & \(\mathbf{97.32}\,{\scriptstyle \pm 0.18}\) & \(\mathbf{97.65}\,{\scriptstyle \pm 0.22}\) & \(\mathbf{99.10}\,{\scriptstyle \pm 0.09}\) & \(99.86\,{\scriptstyle \pm 0.05}\) & \(\underline{91.52}\,{\scriptstyle \pm 0.09}\) & \(\mathbf{97.46}\) \\
\midrule[\heavyrulewidth]
\multicolumn{10}{c}{\textbf{Precision}} \\
\midrule
 & {\renewcommand{\arraystretch}{0.6}\diagbox[font={\fontsize{5.5}{6.6}\selectfont},width=8.5em,height=1.1em]{Method}{Dataset}} & Adult & Default & Magic & Shoppers & Diabetes & Beijing & News & Average \\
\midrule
 & \textsc{Train Set} & \(99.52\,{\scriptstyle \pm 0.00}\) & \(99.69\,{\scriptstyle \pm 0.00}\) & \(99.54\,{\scriptstyle \pm 0.00}\) & \(98.77\,{\scriptstyle \pm 0.00}\) & \(99.37\,{\scriptstyle \pm 0.00}\) & \(99.42\,{\scriptstyle \pm 0.00}\) & \(99.47\,{\scriptstyle \pm 0.00}\) & \(99.40\) \\
\midrule[0.2pt]
\multirow{7}{*}[0pt]{\rotatebox[origin=c]{90}{\fontsize{5.5}{6.6}\selectfont Single-dataset}} & \textsc{SMOTE} & \(92.63\,{\scriptstyle \pm 0.28}\) & \(97.19\,{\scriptstyle \pm 0.46}\) & \(97.80\,{\scriptstyle \pm 0.12}\) & \(92.60\,{\scriptstyle \pm 0.43}\) & \(87.01\,{\scriptstyle \pm 0.21}\) & \(98.99\,{\scriptstyle \pm 0.22}\) & \(87.44\,{\scriptstyle \pm 0.58}\) & \(93.38\) \\
 & \textsc{CTGAN} & \(77.25\,{\scriptstyle \pm 5.40}\) & \(73.24\,{\scriptstyle \pm 3.44}\) & \(81.36\,{\scriptstyle \pm 3.05}\) & \(83.43\,{\scriptstyle \pm 3.92}\) & \(73.77\,{\scriptstyle \pm 17.87}\) & \(87.49\,{\scriptstyle \pm 2.17}\) & \(91.67\,{\scriptstyle \pm 1.51}\) & \(81.17\) \\
 & \textsc{TVAE} & \(93.25\,{\scriptstyle \pm 2.10}\) & \(86.23\,{\scriptstyle \pm 2.81}\) & \(95.36\,{\scriptstyle \pm 1.69}\) & \(49.53\,{\scriptstyle \pm 1.22}\) & \(39.19\,{\scriptstyle \pm 1.64}\) & \(90.33\,{\scriptstyle \pm 2.83}\) & \(93.85\,{\scriptstyle \pm 1.09}\) & \(78.25\) \\
 & \textsc{GReaT} & \(66.68\,{\scriptstyle \pm 0.37}\) & \(89.89\,{\scriptstyle \pm 0.34}\) & \(84.68\,{\scriptstyle \pm 0.16}\) & \(80.75\,{\scriptstyle \pm 0.92}\) & \(59.51\,{\scriptstyle \pm 0.21}\) & \(96.19\,{\scriptstyle \pm 0.13}\) & -- & \(79.62\) \\
 & \textsc{TabDDPM} & \(95.98\,{\scriptstyle \pm 0.08}\) & \(96.62\,{\scriptstyle \pm 0.37}\) & \(95.98\,{\scriptstyle \pm 0.32}\) & \(91.27\,{\scriptstyle \pm 0.22}\) & \(11.10\,{\scriptstyle \pm 0.03}\) & \(99.20\,{\scriptstyle \pm 0.39}\) & \(0.00\,{\scriptstyle \pm 0.00}\) & \(70.02\) \\
 & \textsc{TabSyn} & \(\underline{99.35}\,{\scriptstyle \pm 0.14}\) & \(\underline{98.59}\,{\scriptstyle \pm 0.07}\) & \(98.61\,{\scriptstyle \pm 0.47}\) & \(\underline{97.90}\,{\scriptstyle \pm 0.40}\) & \(92.27\,{\scriptstyle \pm 0.20}\) & \(98.67\,{\scriptstyle \pm 0.16}\) & \(\underline{96.95}\,{\scriptstyle \pm 0.14}\) & \(\underline{97.48}\) \\
 & \textsc{TabDiff} & \(98.93\,{\scriptstyle \pm 0.31}\) & \(98.56\,{\scriptstyle \pm 0.05}\) & \(\mathbf{99.49}\,{\scriptstyle \pm 0.07}\) & \(97.29\,{\scriptstyle \pm 0.97}\) & \(\underline{93.69}\,{\scriptstyle \pm 0.26}\) & \(\underline{99.39}\,{\scriptstyle \pm 0.12}\) & \(86.73\,{\scriptstyle \pm 0.12}\) & \(96.30\) \\
\midrule[0.2pt]
\multirow{3}{*}[0pt]{\rotatebox[origin=c]{90}{\fontsize{5.5}{6.6}\selectfont Cross-dataset}} & \textsc{TabEBM} & \(24.17\,{\scriptstyle \pm 0.27}\) & \(46.68\,{\scriptstyle \pm 0.46}\) & \(92.89\,{\scriptstyle \pm 0.08}\) & \(47.06\,{\scriptstyle \pm 0.44}\) & \(1.51\,{\scriptstyle \pm 0.35}\) & \(78.75\,{\scriptstyle \pm 0.86}\) & \(38.45\,{\scriptstyle \pm 34.35}\) & \(47.07\) \\
 & \textsc{CTSyn} & \(53.78\,{\scriptstyle \pm 0.25}\) & \(89.97\,{\scriptstyle \pm 0.44}\) & \(87.31\,{\scriptstyle \pm 0.12}\) & \(66.48\,{\scriptstyle \pm 0.30}\) & \(8.32\,{\scriptstyle \pm 0.01}\) & \(83.98\,{\scriptstyle \pm 0.27}\) & \(77.85\,{\scriptstyle \pm 0.06}\) & \(66.81\) \\
 & \textsc{CDMD (Ours)} & \(\mathbf{99.59}\,{\scriptstyle \pm 0.06}\) & \(\mathbf{98.77}\,{\scriptstyle \pm 0.06}\) & \(\underline{99.22}\,{\scriptstyle \pm 0.43}\) & \(\mathbf{98.65}\,{\scriptstyle \pm 0.24}\) & \(\mathbf{99.64}\,{\scriptstyle \pm 0.07}\) & \(\mathbf{99.52}\,{\scriptstyle \pm 0.13}\) & \(\mathbf{98.50}\,{\scriptstyle \pm 0.29}\) & \(\mathbf{99.13}\) \\
\midrule[\heavyrulewidth]
\multicolumn{10}{c}{\textbf{Recall}} \\
\midrule
 & {\renewcommand{\arraystretch}{0.6}\diagbox[font={\fontsize{5.5}{6.6}\selectfont},width=8.5em,height=1.1em]{Method}{Dataset}} & Adult & Default & Magic & Shoppers & Diabetes & Beijing & News & Average \\
\midrule
 & \textsc{Train Set} & \(48.55\,{\scriptstyle \pm 0.00}\) & \(50.39\,{\scriptstyle \pm 0.00}\) & \(50.20\,{\scriptstyle \pm 0.00}\) & \(49.40\,{\scriptstyle \pm 0.00}\) & \(48.94\,{\scriptstyle \pm 0.00}\) & \(50.12\,{\scriptstyle \pm 0.00}\) & \(49.71\,{\scriptstyle \pm 0.00}\) & \(49.62\) \\
\midrule[0.2pt]
\multirow{7}{*}[0pt]{\rotatebox[origin=c]{90}{\fontsize{5.5}{6.6}\selectfont Single-dataset}} & \textsc{SMOTE} & \(41.35\,{\scriptstyle \pm 0.30}\) & \(44.87\,{\scriptstyle \pm 0.04}\) & \(45.70\,{\scriptstyle \pm 0.25}\) & \(41.76\,{\scriptstyle \pm 0.24}\) & \(41.51\,{\scriptstyle \pm 0.04}\) & \(\mathbf{41.43}\,{\scriptstyle \pm 0.29}\) & \(\mathbf{49.56}\,{\scriptstyle \pm 0.24}\) & \(43.74\) \\
 & \textsc{CTGAN} & \(26.70\,{\scriptstyle \pm 1.75}\) & \(18.23\,{\scriptstyle \pm 1.01}\) & \(9.01\,{\scriptstyle \pm 0.73}\) & \(25.62\,{\scriptstyle \pm 0.83}\) & \(12.76\,{\scriptstyle \pm 2.10}\) & \(24.32\,{\scriptstyle \pm 1.11}\) & \(9.75\,{\scriptstyle \pm 0.06}\) & \(18.06\) \\
 & \textsc{TVAE} & \(33.07\,{\scriptstyle \pm 2.48}\) & \(23.35\,{\scriptstyle \pm 1.42}\) & \(28.86\,{\scriptstyle \pm 1.88}\) & \(13.56\,{\scriptstyle \pm 1.31}\) & \(18.24\,{\scriptstyle \pm 2.74}\) & \(24.74\,{\scriptstyle \pm 1.67}\) & \(26.00\,{\scriptstyle \pm 0.80}\) & \(23.97\) \\
 & \textsc{GReaT} & \(\mathbf{49.31}\,{\scriptstyle \pm 0.08}\) & \(42.88\,{\scriptstyle \pm 0.55}\) & \(39.10\,{\scriptstyle \pm 0.07}\) & \(45.50\,{\scriptstyle \pm 0.21}\) & \(\mathbf{54.44}\,{\scriptstyle \pm 0.08}\) & \(\underline{36.13}\,{\scriptstyle \pm 0.11}\) & -- & \(\mathbf{44.56}\) \\
 & \textsc{TabDDPM} & \(46.90\,{\scriptstyle \pm 0.23}\) & \(45.04\,{\scriptstyle \pm 0.97}\) & \(\mathbf{48.13}\,{\scriptstyle \pm 0.30}\) & \(\mathbf{48.16}\,{\scriptstyle \pm 0.64}\) & \(0.04\,{\scriptstyle \pm 0.01}\) & \(34.93\,{\scriptstyle \pm 0.23}\) & \(0.00\,{\scriptstyle \pm 0.00}\) & \(31.88\) \\
 & \textsc{TabSyn} & \(\underline{46.98}\,{\scriptstyle \pm 0.10}\) & \(46.14\,{\scriptstyle \pm 0.22}\) & \(\underline{48.06}\,{\scriptstyle \pm 0.29}\) & \(45.78\,{\scriptstyle \pm 0.83}\) & \(34.93\,{\scriptstyle \pm 0.15}\) & \(35.40\,{\scriptstyle \pm 0.27}\) & \(36.20\,{\scriptstyle \pm 0.15}\) & \(41.93\) \\
 & \textsc{TabDiff} & \(46.39\,{\scriptstyle \pm 0.24}\) & \(\underline{47.50}\,{\scriptstyle \pm 0.30}\) & \(46.83\,{\scriptstyle \pm 0.67}\) & \(45.03\,{\scriptstyle \pm 0.22}\) & \(\underline{45.79}\,{\scriptstyle \pm 0.03}\) & \(35.44\,{\scriptstyle \pm 0.35}\) & \(26.94\,{\scriptstyle \pm 0.24}\) & \(41.99\) \\
\midrule[0.2pt]
\multirow{3}{*}[0pt]{\rotatebox[origin=c]{90}{\fontsize{5.5}{6.6}\selectfont Cross-dataset}} & \textsc{TabEBM} & \(1.32\,{\scriptstyle \pm 0.02}\) & \(6.41\,{\scriptstyle \pm 0.19}\) & \(34.35\,{\scriptstyle \pm 0.05}\) & \(13.41\,{\scriptstyle \pm 0.07}\) & \(0.00\,{\scriptstyle \pm 0.00}\) & \(14.87\,{\scriptstyle \pm 0.25}\) & \(5.87\,{\scriptstyle \pm 7.93}\) & \(10.89\) \\
 & \textsc{CTSyn} & \(17.57\,{\scriptstyle \pm 0.18}\) & \(18.80\,{\scriptstyle \pm 0.08}\) & \(11.19\,{\scriptstyle \pm 0.12}\) & \(25.36\,{\scriptstyle \pm 0.28}\) & \(3.18\,{\scriptstyle \pm 0.01}\) & \(4.78\,{\scriptstyle \pm 0.12}\) & \(27.76\,{\scriptstyle \pm 0.23}\) & \(15.52\) \\
 & \textsc{CDMD (Ours)} & \(46.93\,{\scriptstyle \pm 0.14}\) & \(\mathbf{48.15}\,{\scriptstyle \pm 0.33}\) & \(47.41\,{\scriptstyle \pm 0.23}\) & \(\underline{46.77}\,{\scriptstyle \pm 0.04}\) & \(44.13\,{\scriptstyle \pm 0.14}\) & \(34.98\,{\scriptstyle \pm 0.14}\) & \(\underline{41.36}\,{\scriptstyle \pm 0.03}\) & \(\underline{44.25}\) \\
\bottomrule
\end{tabular}
}%
\end{table}

\begin{table}[t]
\caption{In-domain cross-dataset generation results for Privacy.}
\label{tab:dcr_score__authenticity}
\centering
\small
\resizebox{\linewidth}{!}{%
\begin{tabular}{@{}c@{\hspace{0.4em}}l|c|c|c|c|c|c|c|c}
\toprule
\multicolumn{10}{c}{\textbf{DCR}} \\
\midrule
 & {\renewcommand{\arraystretch}{0.6}\diagbox[font={\fontsize{5.5}{6.6}\selectfont},width=8.5em,height=1.1em]{Method}{Dataset}} & Adult & Default & Magic & Shoppers & Diabetes & Beijing & News & Average \\
\midrule
 & \textsc{Train Set} & \(0.19\,{\scriptstyle \pm 0.00}\) & \(0.21\,{\scriptstyle \pm 0.00}\) & \(1.20\,{\scriptstyle \pm 0.00}\) & \(2.27\,{\scriptstyle \pm 0.00}\) & \(0.01\,{\scriptstyle \pm 0.00}\) & \(0.03\,{\scriptstyle \pm 0.00}\) & \(0.00\,{\scriptstyle \pm 0.00}\) & \(0.56\) \\
\midrule[0.2pt]
\multirow{7}{*}[0pt]{\rotatebox[origin=c]{90}{\fontsize{5.5}{6.6}\selectfont Single-dataset}} & \textsc{SMOTE} & \(18.23\,{\scriptstyle \pm 0.12}\) & \(16.94\,{\scriptstyle \pm 0.13}\) & \(8.76\,{\scriptstyle \pm 0.29}\) & \(7.66\,{\scriptstyle \pm 0.76}\) & \(1.11\,{\scriptstyle \pm 0.09}\) & \(23.26\,{\scriptstyle \pm 0.27}\) & \(3.52\,{\scriptstyle \pm 0.18}\) & \(11.35\) \\
 & \textsc{CTGAN} & \(98.31\,{\scriptstyle \pm 0.20}\) & \(97.46\,{\scriptstyle \pm 1.01}\) & \(98.79\,{\scriptstyle \pm 0.44}\) & \(\underline{97.81}\,{\scriptstyle \pm 0.66}\) & \(\mathbf{100.00}\,{\scriptstyle \pm 0.00}\) & \(\underline{99.94}\,{\scriptstyle \pm 0.09}\) & \(96.98\,{\scriptstyle \pm 0.78}\) & \(\underline{98.47}\) \\
 & \textsc{TVAE} & \(\underline{98.58}\,{\scriptstyle \pm 0.14}\) & \(\mathbf{98.77}\,{\scriptstyle \pm 0.90}\) & \(\underline{99.08}\,{\scriptstyle \pm 1.06}\) & \(97.08\,{\scriptstyle \pm 2.78}\) & \(98.71\,{\scriptstyle \pm 0.68}\) & \(98.09\,{\scriptstyle \pm 0.98}\) & \(96.27\,{\scriptstyle \pm 0.80}\) & \(98.09\) \\
 & \textsc{GReaT} & \(95.15\,{\scriptstyle \pm 0.23}\) & \(98.12\,{\scriptstyle \pm 0.72}\) & \(97.26\,{\scriptstyle \pm 0.30}\) & \(\mathbf{99.75}\,{\scriptstyle \pm 0.34}\) & \(97.97\,{\scriptstyle \pm 0.43}\) & \(87.16\,{\scriptstyle \pm 0.31}\) & -- & \(95.90\) \\
 & \textsc{TabDDPM} & \(97.76\,{\scriptstyle \pm 0.32}\) & \(98.22\,{\scriptstyle \pm 0.23}\) & \(94.31\,{\scriptstyle \pm 0.17}\) & \(83.42\,{\scriptstyle \pm 0.62}\) & \(98.77\,{\scriptstyle \pm 0.34}\) & \(99.31\,{\scriptstyle \pm 0.45}\) & \(\mathbf{100.00}\,{\scriptstyle \pm 0.00}\) & \(95.97\) \\
 & \textsc{TabSyn} & \(96.75\,{\scriptstyle \pm 0.25}\) & \(97.97\,{\scriptstyle \pm 0.38}\) & \(96.80\,{\scriptstyle \pm 0.29}\) & \(97.33\,{\scriptstyle \pm 1.58}\) & \(\underline{99.72}\,{\scriptstyle \pm 0.39}\) & \(98.43\,{\scriptstyle \pm 0.41}\) & \(\underline{98.22}\,{\scriptstyle \pm 0.34}\) & \(97.89\) \\
 & \textsc{TabDiff} & \(85.55\,{\scriptstyle \pm 0.20}\) & \(85.69\,{\scriptstyle \pm 0.57}\) & \(95.46\,{\scriptstyle \pm 0.21}\) & \(97.11\,{\scriptstyle \pm 0.20}\) & \(86.15\,{\scriptstyle \pm 0.51}\) & \(98.47\,{\scriptstyle \pm 0.41}\) & \(91.89\,{\scriptstyle \pm 0.11}\) & \(91.47\) \\
\midrule[0.2pt]
\multirow{3}{*}[0pt]{\rotatebox[origin=c]{90}{\fontsize{5.5}{6.6}\selectfont Cross-dataset}} & \textsc{TabEBM} & \(82.08\,{\scriptstyle \pm 0.66}\) & \(95.01\,{\scriptstyle \pm 0.29}\) & \(40.14\,{\scriptstyle \pm 0.23}\) & \(80.30\,{\scriptstyle \pm 0.61}\) & \(86.89\,{\scriptstyle \pm 0.10}\) & \(87.89\,{\scriptstyle \pm 0.54}\) & \(48.93\,{\scriptstyle \pm 0.61}\) & \(74.46\) \\
 & \textsc{CTSyn} & \(\mathbf{100.00}\,{\scriptstyle \pm 0.00}\) & \(\underline{98.51}\,{\scriptstyle \pm 0.57}\) & \(\mathbf{100.00}\,{\scriptstyle \pm 0.00}\) & \(95.81\,{\scriptstyle \pm 1.78}\) & \(\mathbf{100.00}\,{\scriptstyle \pm 0.00}\) & \(\mathbf{100.00}\,{\scriptstyle \pm 0.00}\) & \(96.72\,{\scriptstyle \pm 0.88}\) & \(\mathbf{98.72}\) \\
 & \textsc{CDMD (Ours)} & \(97.25\,{\scriptstyle \pm 0.64}\) & \(97.79\,{\scriptstyle \pm 1.04}\) & \(96.52\,{\scriptstyle \pm 1.42}\) & \(95.50\,{\scriptstyle \pm 0.48}\) & \(98.53\,{\scriptstyle \pm 0.27}\) & \(98.13\,{\scriptstyle \pm 0.30}\) & \(97.04\,{\scriptstyle \pm 0.17}\) & \(97.25\) \\
\midrule[\heavyrulewidth]
\multicolumn{10}{c}{\textbf{Authenticity}} \\
\midrule
 & {\renewcommand{\arraystretch}{0.6}\diagbox[font={\fontsize{5.5}{6.6}\selectfont},width=8.5em,height=1.1em]{Method}{Dataset}} & Adult & Default & Magic & Shoppers & Diabetes & Beijing & News & Average \\
\midrule
 & \textsc{Train Set} & \(0.00\,{\scriptstyle \pm 0.00}\) & \(0.00\,{\scriptstyle \pm 0.00}\) & \(0.00\,{\scriptstyle \pm 0.00}\) & \(0.00\,{\scriptstyle \pm 0.00}\) & \(0.00\,{\scriptstyle \pm 0.00}\) & \(0.00\,{\scriptstyle \pm 0.00}\) & \(0.00\,{\scriptstyle \pm 0.00}\) & \(0.00\) \\
\midrule[0.2pt]
\multirow{7}{*}[0pt]{\rotatebox[origin=c]{90}{\fontsize{5.5}{6.6}\selectfont Single-dataset}} & \textsc{SMOTE} & \(28.87\,{\scriptstyle \pm 0.40}\) & \(30.49\,{\scriptstyle \pm 0.23}\) & \(25.21\,{\scriptstyle \pm 0.36}\) & \(26.20\,{\scriptstyle \pm 0.56}\) & \(24.04\,{\scriptstyle \pm 0.11}\) & \(29.58\,{\scriptstyle \pm 0.14}\) & \(25.18\,{\scriptstyle \pm 0.20}\) & \(27.08\) \\
 & \textsc{CTGAN} & \(59.51\,{\scriptstyle \pm 0.46}\) & \(62.05\,{\scriptstyle \pm 0.49}\) & \(\underline{76.55}\,{\scriptstyle \pm 0.26}\) & \(62.81\,{\scriptstyle \pm 0.20}\) & \(73.42\,{\scriptstyle \pm 2.02}\) & \(72.01\,{\scriptstyle \pm 0.64}\) & \(\underline{74.04}\,{\scriptstyle \pm 0.60}\) & \(68.63\) \\
 & \textsc{TVAE} & \(56.33\,{\scriptstyle \pm 0.91}\) & \(\underline{64.75}\,{\scriptstyle \pm 1.49}\) & \(63.73\,{\scriptstyle \pm 1.58}\) & \(\mathbf{77.02}\,{\scriptstyle \pm 1.29}\) & \(65.99\,{\scriptstyle \pm 1.70}\) & \(71.07\,{\scriptstyle \pm 2.27}\) & \(66.32\,{\scriptstyle \pm 1.19}\) & \(66.46\) \\
 & \textsc{GReaT} & \(51.24\,{\scriptstyle \pm 0.12}\) & \(52.14\,{\scriptstyle \pm 0.39}\) & \(56.42\,{\scriptstyle \pm 0.80}\) & \(52.33\,{\scriptstyle \pm 0.63}\) & \(50.45\,{\scriptstyle \pm 0.23}\) & \(59.34\,{\scriptstyle \pm 0.22}\) & -- & \(53.65\) \\
 & \textsc{TabDDPM} & \(51.26\,{\scriptstyle \pm 0.38}\) & \(51.14\,{\scriptstyle \pm 0.28}\) & \(51.45\,{\scriptstyle \pm 0.53}\) & \(46.81\,{\scriptstyle \pm 0.50}\) & \(\underline{98.53}\,{\scriptstyle \pm 0.15}\) & \(63.25\,{\scriptstyle \pm 0.08}\) & \(\mathbf{100.00}\,{\scriptstyle \pm 0.00}\) & \(66.06\) \\
 & \textsc{TabSyn} & \(51.20\,{\scriptstyle \pm 0.34}\) & \(50.59\,{\scriptstyle \pm 0.20}\) & \(51.74\,{\scriptstyle \pm 0.62}\) & \(51.08\,{\scriptstyle \pm 0.24}\) & \(57.23\,{\scriptstyle \pm 0.09}\) & \(62.78\,{\scriptstyle \pm 0.26}\) & \(56.61\,{\scriptstyle \pm 0.17}\) & \(54.46\) \\
 & \textsc{TabDiff} & \(48.22\,{\scriptstyle \pm 0.27}\) & \(48.14\,{\scriptstyle \pm 0.18}\) & \(51.76\,{\scriptstyle \pm 0.67}\) & \(51.75\,{\scriptstyle \pm 0.61}\) & \(49.12\,{\scriptstyle \pm 0.22}\) & \(62.69\,{\scriptstyle \pm 0.20}\) & \(57.68\,{\scriptstyle \pm 0.14}\) & \(52.77\) \\
\midrule[0.2pt]
\multirow{3}{*}[0pt]{\rotatebox[origin=c]{90}{\fontsize{5.5}{6.6}\selectfont Cross-dataset}} & \textsc{TabEBM} & \(\mathbf{85.62}\,{\scriptstyle \pm 1.17}\) & \(\mathbf{83.29}\,{\scriptstyle \pm 1.06}\) & \(41.04\,{\scriptstyle \pm 0.19}\) & \(\underline{71.44}\,{\scriptstyle \pm 1.07}\) & \(\mathbf{99.97}\,{\scriptstyle \pm 0.01}\) & \(\underline{76.71}\,{\scriptstyle \pm 0.32}\) & \(58.31\,{\scriptstyle \pm 0.10}\) & \(\underline{73.77}\) \\
 & \textsc{CTSyn} & \(\underline{68.95}\,{\scriptstyle \pm 0.13}\) & \(64.60\,{\scriptstyle \pm 1.03}\) & \(\mathbf{78.58}\,{\scriptstyle \pm 1.44}\) & \(66.73\,{\scriptstyle \pm 0.51}\) & \(84.70\,{\scriptstyle \pm 0.11}\) & \(\mathbf{92.92}\,{\scriptstyle \pm 0.13}\) & \(62.26\,{\scriptstyle \pm 0.15}\) & \(\mathbf{74.11}\) \\
 & \textsc{CDMD (Ours)} & \(51.00\,{\scriptstyle \pm 0.39}\) & \(50.21\,{\scriptstyle \pm 0.55}\) & \(51.75\,{\scriptstyle \pm 0.55}\) & \(49.30\,{\scriptstyle \pm 0.09}\) & \(52.14\,{\scriptstyle \pm 0.20}\) & \(62.77\,{\scriptstyle \pm 0.30}\) & \(53.66\,{\scriptstyle \pm 0.25}\) & \(52.97\) \\
\bottomrule
\end{tabular}
}%
\end{table}

\subsection{Few-Shot Generation}
\label{app:few_shot_generation}

\begin{figure}[t]
    \centering
    \includegraphics[width=\linewidth]{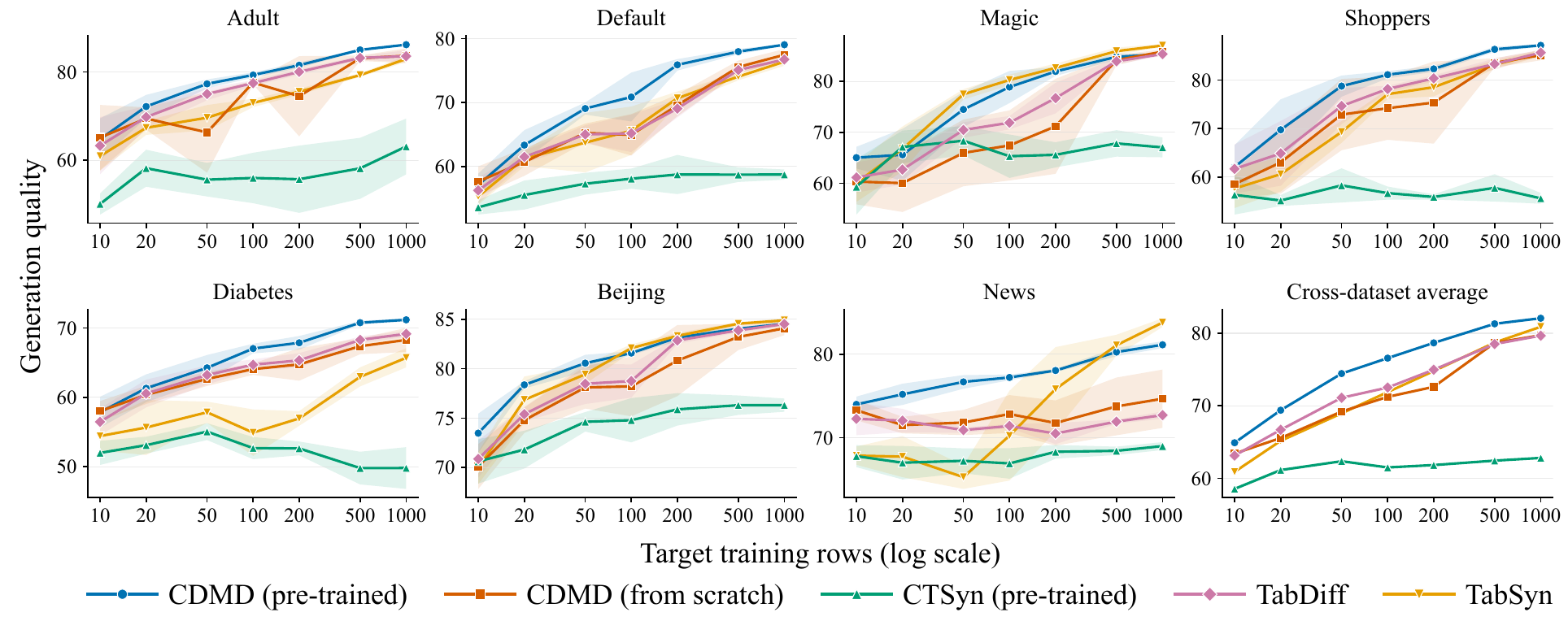}
    \caption{Few-shot performance across seven datasets and their
    cross-dataset average. Curves show mean performance, and shaded
    regions indicate one standard deviation across seeds.}
    \label{app:fig:few-shot-performance}
\end{figure}

For the few-shot generation experiments (Section~\ref{sec:exp:constrained_generation}), we complement the results presented in the main paper with the generation quality score in Figure~\ref{app:fig:few-shot-performance}.

\subsection{Few-Epoch Generation}
\label{app:few_epoch_generation}

\begin{figure}[t]
    \centering
    \includegraphics[width=\linewidth]{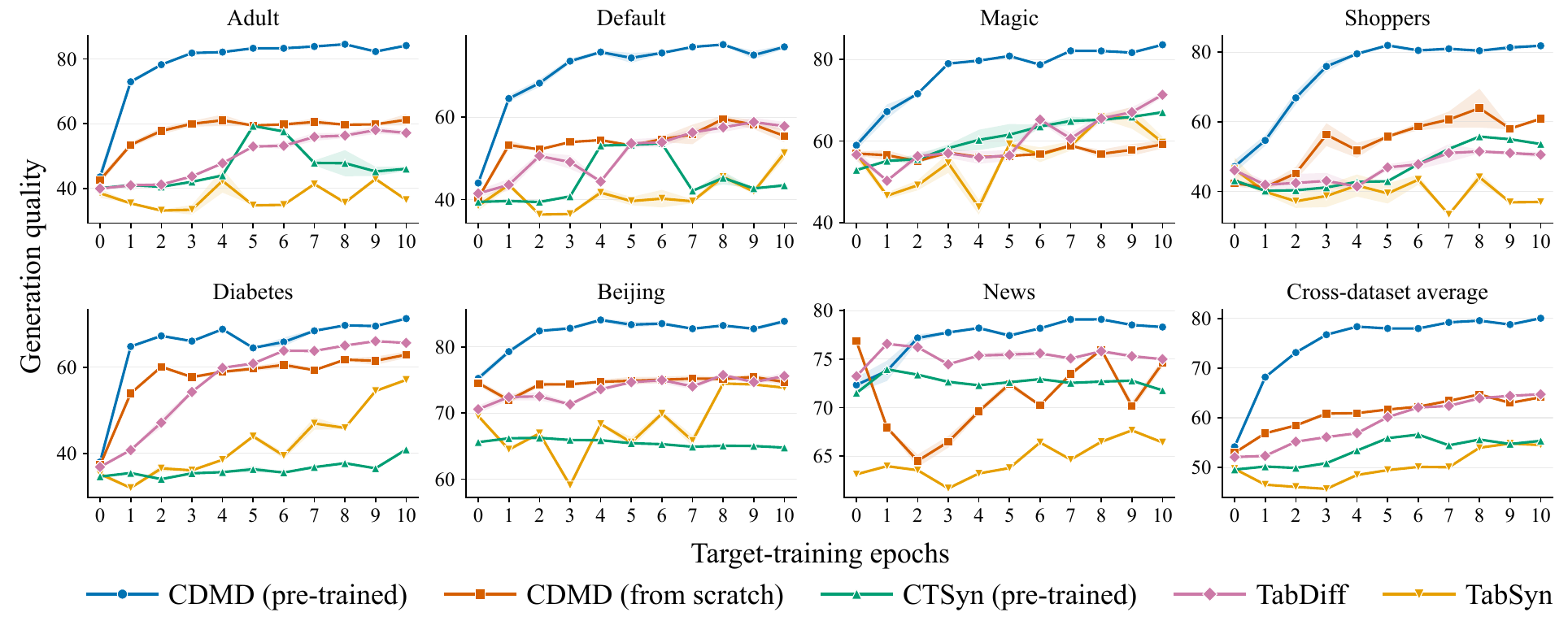}
    \caption{Few-epoch performance across seven datasets and their
    cross-dataset average. Curves show mean performance, and shaded
    regions indicate one standard deviation across seeds.}
    \label{app:fig:few-epoch-performance}
\end{figure}

For the few-epoch generation experiments (Section~\ref{sec:exp:constrained_generation}), we plot the generation quality score against the number of training epochs in Figure~\ref{app:fig:few-epoch-performance}.

\end{document}